\documentclass[]{hcrg-nasa}
\graphicspath{{figures/}}
\newif\ifauthorcomments
\authorcommentstrue
\newcommand{\authorcomment}[2][]{\ifauthorcomments\par\begingroup\small\interlinepenalty=10000\noindent\if\relax\detokenize{#1}\relax\color{red}\textbf{Author comment:}\else\color{blue}\textbf{@#1:}\fi\ #2\par\endgroup\fi}

\abstract{
Visual captures provide detailed scene appearance, but converting them into interactive simulation environments requires object reconstruction, physical modeling, and correction of intermediate errors. We present SceneAgent, an agentic pipeline that coordinates these steps through simulator-based visual inspection and repair. The pipeline processes 3D Gaussian Splatting (3DGS), photogrammetry, LiDAR captures, AI-generated 3DGS scenes, and 4D videos of dynamic scenes, and combines object segmentation, physical parameter estimation, part decomposition, articulation, and deformable simulation. 
To support policy learning, SceneAgent generates diverse object variants (``digital sisters'') and scripted demonstrations. On a cluttered bin scene from GraspClutter6D, SceneAgent raises F1 at 1~cm for every coding agent with scorable outputs both with and without its skills, and its reconstructions are better than SimFoundry~\cite{ranawaka2026simfoundry}: all agent configurations exceed SimFoundry's F1 score, with a best of 60.75\% versus 6.9\%. On three dynamic scenes from 4DCodeBench~\cite{shen20264dcodebench}, SceneAgent raises the mean dynamic IoU of Gemini 4 Argon from 0.86\% to 99.09\%. We further demonstrate diverse simulated interactions and preliminary policy transfer to a real robot.
}
\keywords{SceneAgent, real-to-sim, sim-to-real, 3D Gaussian Splatting, 4D reconstruction, simulation environments, predictive physics, robot policy evaluation, policy training, digital sisters}
\website{https://computationalrobotics.seas.harvard.edu/SceneAgent/}{\texttt{https://computationalrobotics.seas.harvard.edu/SceneAgent/}}

\hypersetup{
  pdftitle={SceneAgent: An Agentic Pipeline from 3D and 4D Captures to Simulation Environments},
  pdfauthor={Luke Hollis, Tianxing Fan, Heng Yang},
  pdfsubject={SceneAgent: 3D and 4D captures to simulation environments with predictive physics},
  pdfkeywords={SceneAgent; real-to-sim; sim-to-real; 3D Gaussian Splatting; 4D reconstruction; simulation environments; predictive physics; robot policy evaluation; policy training; digital sisters},
}

\begin{document}

\title{SceneAgent: An Agentic Pipeline from 3D and 4D Captures to Simulation Environments}

\author[1*]{Luke Hollis}
\author[1*]{Tianxing Fan}
\author[1]{Heng Yang}
\affiliation[1]{Computational Robotics Group, School of Engineering and Applied Sciences, Harvard University}
\contribution[*]{Equal contribution}
\contribution[]{\email{lhollis@g.harvard.edu}, \email{tianxing\_fan@seas.harvard.edu}, \email{hankyang@seas.harvard.edu}}

\maketitle

\begingroup
\setlength{\parskip}{0pt}
\vfill
\nointerlineskip
\noindent\begin{minipage}{\linewidth}
\centering
\begin{minipage}{\linewidth}
\includegraphics[width=0.325\linewidth,viewport={0 0 114.573 64.446},clip]{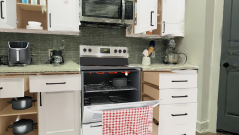}\hfill
\includegraphics[width=0.325\linewidth,viewport={118.555 0 233.122 64.446},clip]{fig_teaser}\hfill
\includegraphics[width=0.325\linewidth,viewport={237.111 0 351.684 64.446},clip]{fig_teaser}
\par\smallskip
\includegraphics[width=0.325\linewidth,viewport={355.666 0 470.237 64.446},clip]{fig_teaser}\hfill
\includegraphics[width=0.325\linewidth]{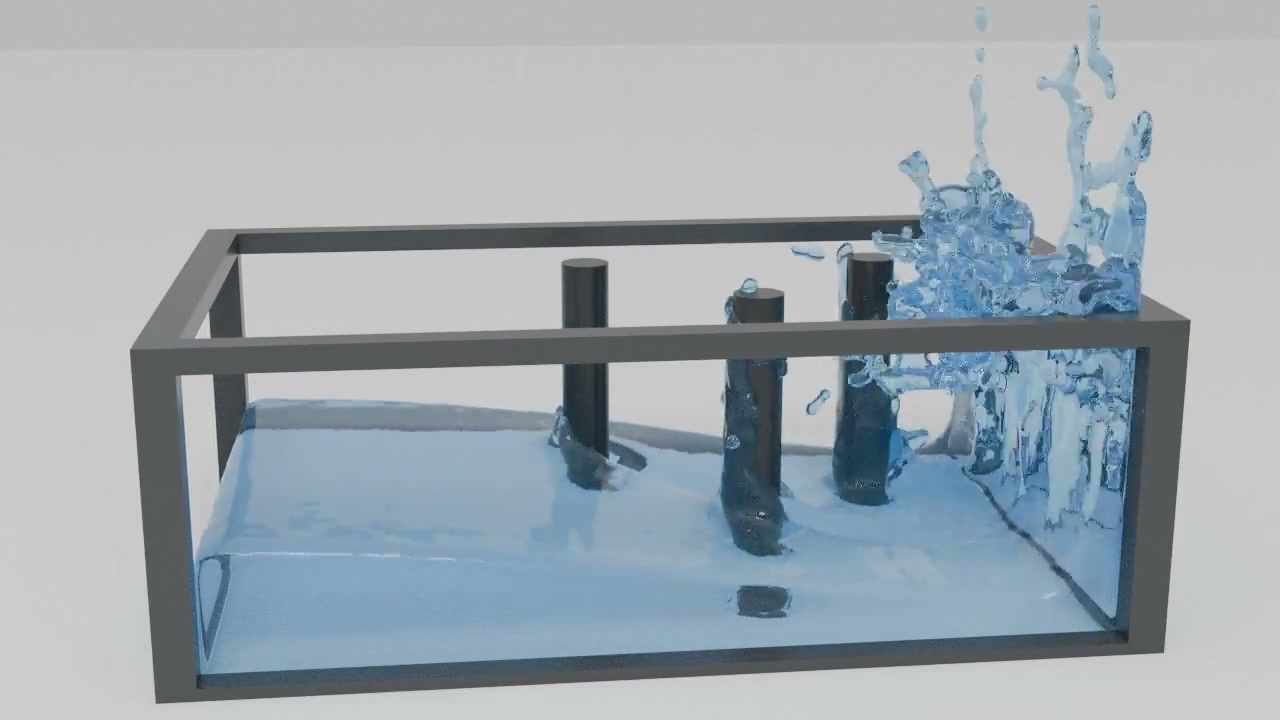}\hfill
\includegraphics[width=0.325\linewidth]{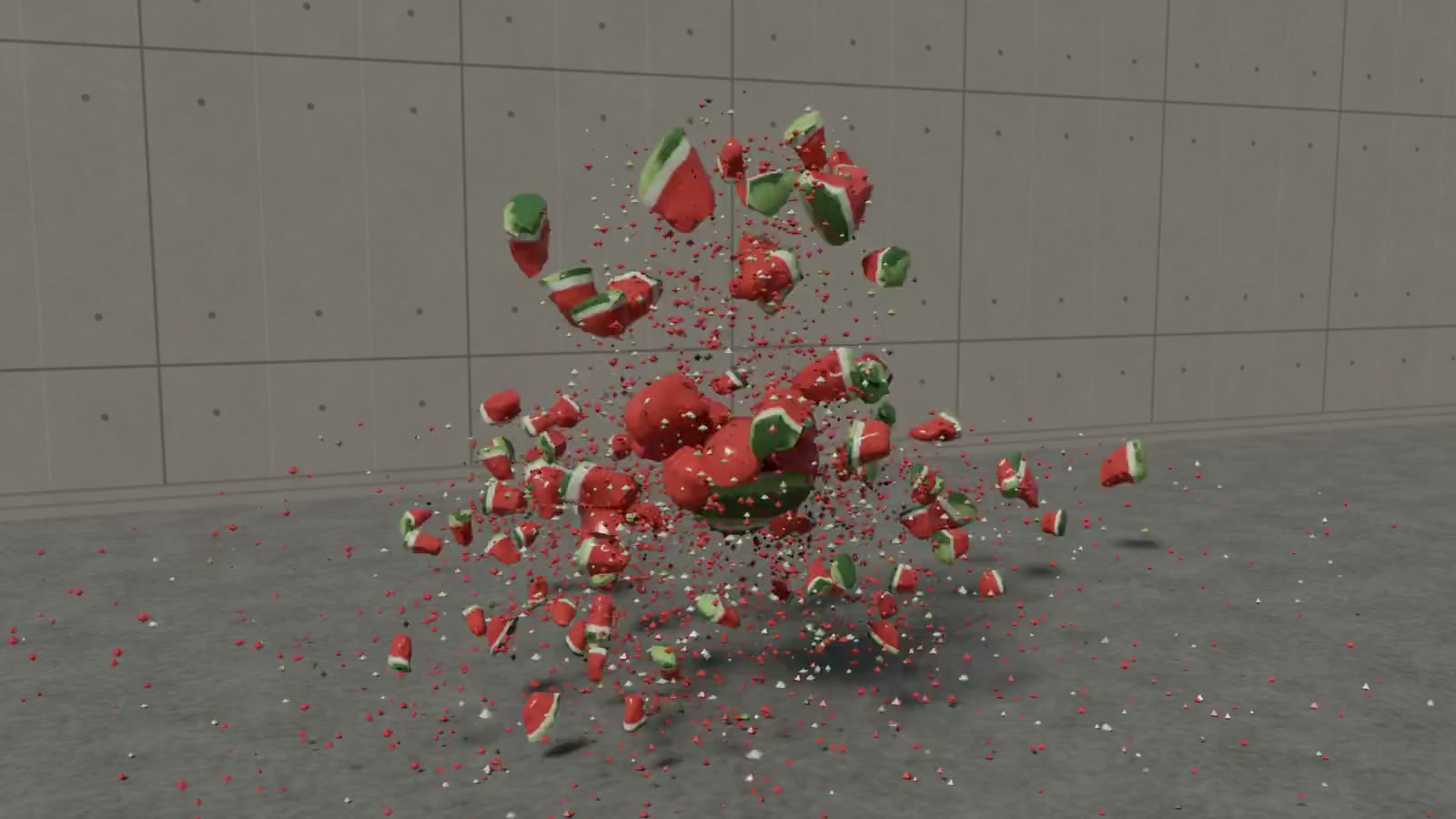}
\end{minipage}
\captionof{figure}{Simulation scenes and tasks from SceneAgent (left to right, top to bottom): a kitchen from CollectData with its doors, drawers, and appliances articulated; picking and plugging cables into a data center server rack; Sharpa hands putting sunglasses on a statue; a TeleAvatar with two Sharpa hands at a clamp in a lab space; and SceneAgent's 4DCodeBench reconstructions of a liquid dam break around rigid cylindrical pillars (B\_09) and a layered watermelon breaking apart on impact (Z\_12).}
\label{fig:teaser}
\end{minipage}\par
\endgroup
\clearpage

\section{Introduction}
\label{sec:introduction}

Generalist robot policies~\cite{brohan2023rt2,octo2024octo,kim2024openvla,black2024pi0,intelligence2025pi05,al2025gr00t} learn from large teleoperated datasets to perform a wide range of manipulation tasks specified by language instructions. Collecting these datasets is labor-intensive and can take months or years. Similarly, evaluating the resulting policies in a target environment requires physical robot trials, which are time-consuming, difficult to reproduce, and costly to scale~\cite{atreya2025roboarena}.

Simulation offers a way to scale both data collection and evaluation. Reconstructed environments can generate demonstrations~\cite{yu2025real2render2real,yang2025demonstration,wang2026exogs}, support policies that transfer to physical robots~\cite{qureshi2024splatsim,li2024robogsim,zhang2026gase}, and provide estimates of real-world policy performance~\cite{jain2025polaris,ranawaka2026simfoundry,zhang2025realtosim,luo2025sim2val,luo2026x4val}. In particular, 3D Gaussian Splatting (3DGS)~\cite{kerbl2023gaussian} reconstructs photorealistic scene appearance from images, enabling simulators to render observations grounded in a captured workspace~\cite{qureshi2024splatsim,escontrela2025gaussgym,jia2026gsplayground}.

Visual reconstruction alone, however, does not provide the object geometry, physical parameters, or articulation needed for interaction. Real-to-sim systems recover these components by combining segmentation, mesh generation, pose alignment, and physical modeling~\cite{jain2025polaris,ranawaka2026simfoundry,zhang2026gase,gao2026mirage2matter}. SimFoundry, for example, supports articulated objects and estimates their physical parameters~\cite{ranawaka2026simfoundry}. A remaining challenge is to coordinate these components across diverse inputs and correct errors that arise between stages. An incorrect mask, scale, or transform can produce an unusable object even when the individual models return plausible outputs.

\begin{figure}[t]
\centering
\makebox[\linewidth][c]{\includegraphics[width=\dimexpr\linewidth+2cm\relax]{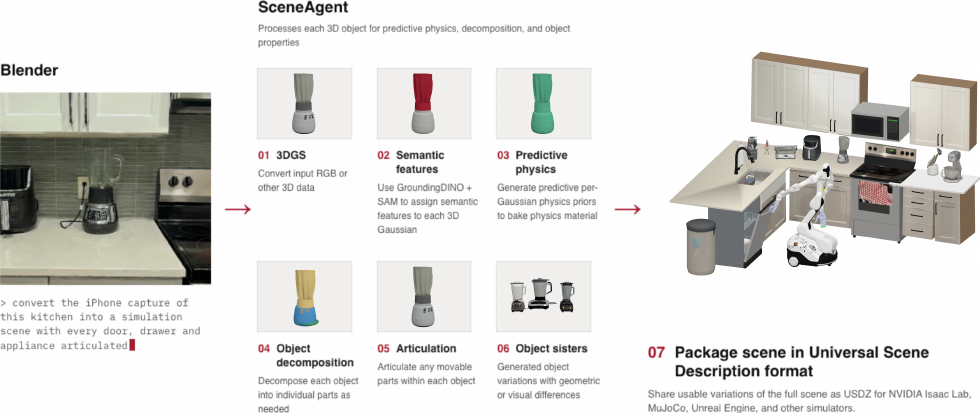}}
\caption{SceneAgent pipeline. Agents coordinate reconstruction, segmentation, physical modeling, and scene export, using simulator renders to inspect and repair intermediate outputs. Inputs include RGB image sequences, 3DGS and photogrammetry assets, AI-generated 3DGS scenes, and LiDAR scans. Section~\ref{sec:4d} extends the pipeline to videos of dynamic scenes.}
\label{fig:pipeline}
\end{figure}

We present SceneAgent, an agentic pipeline that converts 3D and 4D captures into interactive simulation environments (Figure~\ref{fig:pipeline}). Agents execute reconstruction and physical modeling tools, inspect rendered outputs in the target simulator, and repair visible errors before the scene is used downstream. The resulting environments contain segmented objects with estimated physical parameters, decomposed parts, and articulation or soft-body behavior where needed. For 4D captures, which are videos of dynamic scenes, agents also fit simulated dynamics such as liquid flow, deformation, and destruction mechanics to the observed scene (Section~\ref{sec:4d}). SceneAgent also generates digital sisters---related object variants---and combines captured objects with artist-authored environments to expand the range of training scenes.

\begin{figure}[t]
\centering
\includegraphics[width=\linewidth]{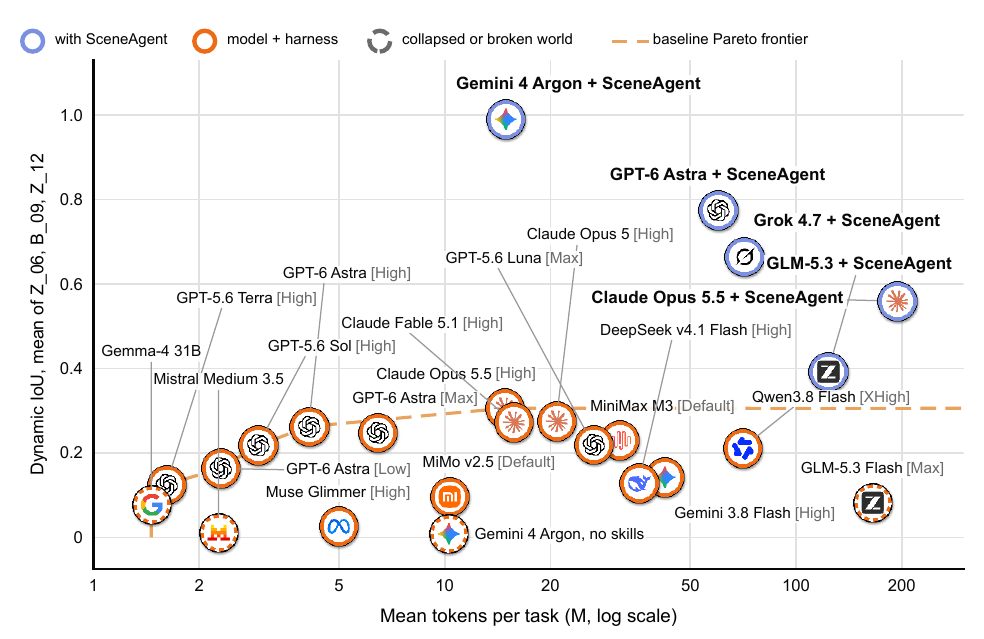}
\caption{4DCodeBench dynamic IoU against tokens per task on the synthetic tasks Z\_06, B\_09 and Z\_12. Baseline scores come from the supplied 4DCodeBench reference table. Baseline tokens are the published averages over 100 tasks.}
\label{fig:4dcodebench-tokens}
\end{figure}

\begin{figure}[t]
\centering
\includegraphics[width=\linewidth]{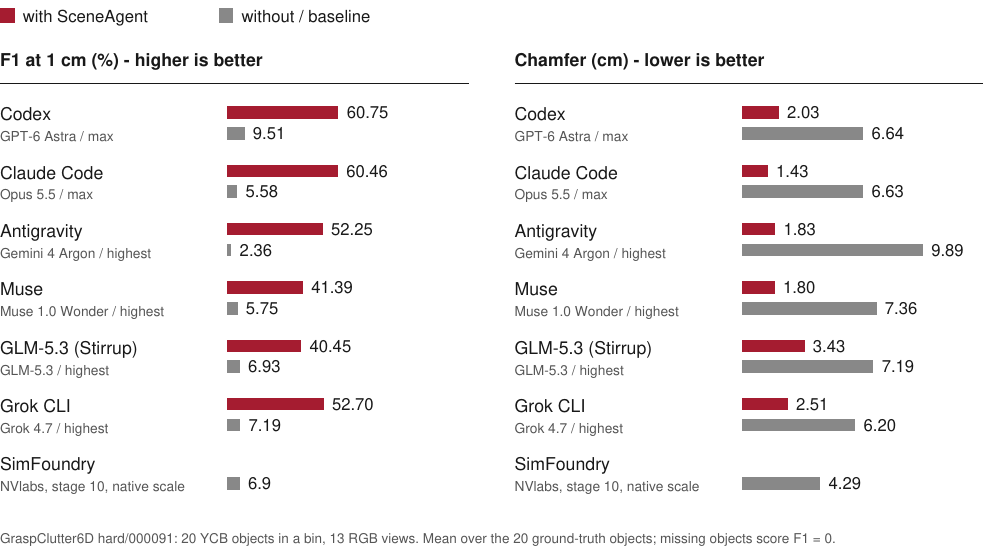}
\caption{Reconstruction of a cluttered bin scene from GraspClutter6D (hard/000091, 20 YCB objects) with and without SceneAgent skills, and with SimFoundry. Higher F1 at 1~cm and lower Chamfer distance indicate closer agreement with the dataset's ground-truth object meshes.}
\label{fig:harness}
\end{figure}

We evaluate reconstruction quality on a cluttered bin scene from GraspClutter6D~\cite{back2025graspclutter6d} and a Messy Kitchen dataset from CollectData~\cite{collectdataio_tidyup} using coding agents with and without SceneAgent, and compare with SimFoundry. Here, a \emph{harness} is the software environment in which a coding agent runs, and \emph{skills} are the reusable instructions and tools that guide its conversion workflow. With SceneAgent, F1 at 1~cm improves for every harness with scorable paired outputs, and reconstruction quality is much better than SimFoundry's (Figure~\ref{fig:harness}). On three dynamic scenes from 4DCodeBench~\cite{shen20264dcodebench}, where agents reconstruct simulations from videos, SceneAgent raises the mean dynamic IoU of Gemini 4 Argon from 0.86\% to 99.09\% (Figure~\ref{fig:4dcodebench-tokens}). We also demonstrate simulated interactions and preliminary real-world deployment after policy learning.

\begin{figure}[t]
\centering
\includegraphics[width=\linewidth]{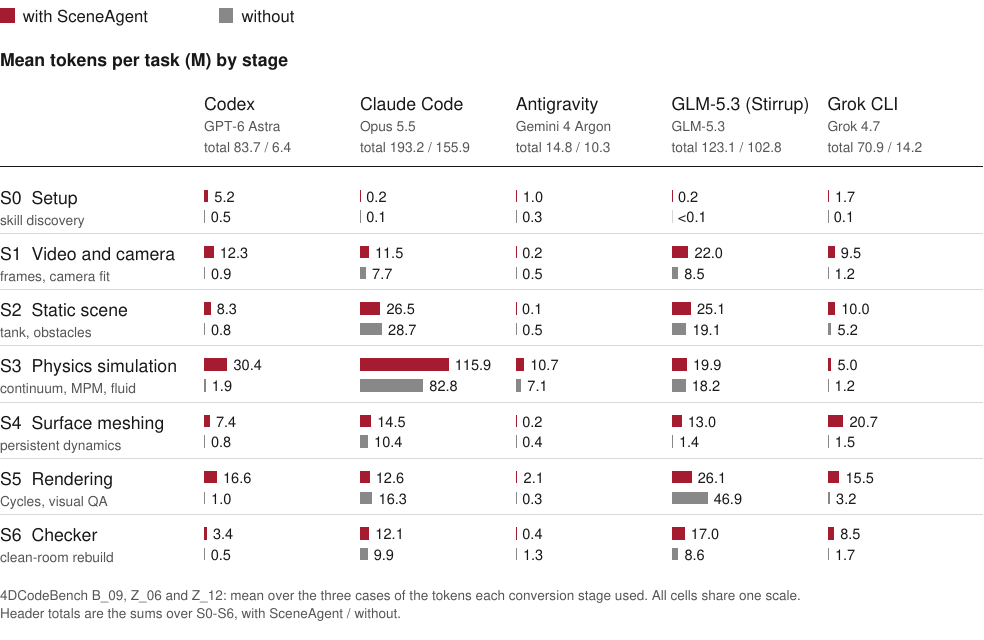}
\caption{4DCodeBench mean tokens per task by conversion stage, with and without SceneAgent skills, on the synthetic tasks Z\_06, B\_09 and Z\_12.}
\label{fig:4dcodebench-stage-tokens}
\end{figure}

\begin{table}[t]
\caption{System comparison. Capabilities of SceneAgent and recent real-to-sim systems for policy learning and evaluation.}
\label{tab:system-comparison}
\makebox[\linewidth][c]{\includegraphics{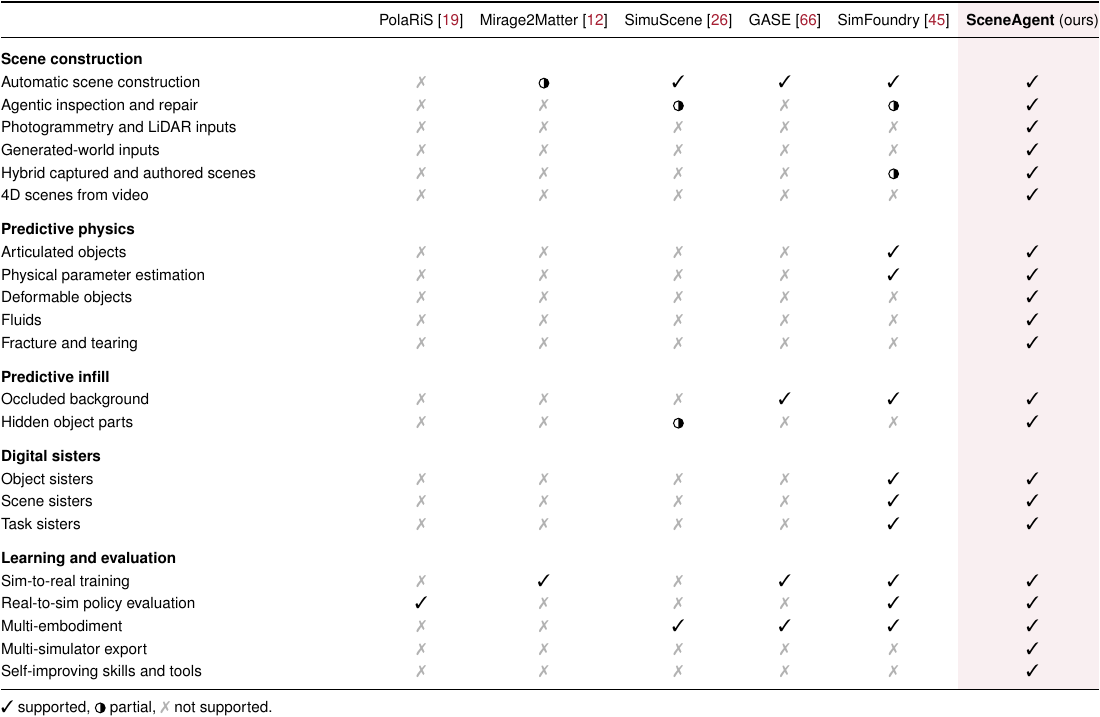}}
\end{table}

In summary, our contributions are:
\begin{itemize}[leftmargin=1.2em, itemsep=2pt, topsep=2pt]
\item An agentic scene-conversion workflow that integrates reconstruction, physical parameter estimation, part decomposition, and articulation with simulator-based visual inspection and repair, and that extends from 3D captures to 4D captures by fitting simulated dynamics to the motion observed in a video.
\item Scene augmentation through digital sisters and hybrid environments that combine captured objects with artist-authored assets, together with an automated research loop that turns failures of a policy on the real robot into simulated training data.
\item A quantitative study of reconstruction quality across coding-agent harnesses on static captures and on dynamic scenes from 4DCodeBench, complemented by qualitative examples of articulated, deformable, and contact-rich interactions and preliminary real-world policy transfer.
\end{itemize}

\section{Related Work}
\label{sec:related}

\paragraph{Real-to-sim reconstruction and policy evaluation.}
Reconstruction-based systems combine scene appearance with interactive geometry to support policy evaluation and training. PolaRiS~\cite{jain2025polaris} combines a Gaussian-splat background with generated object meshes and evaluates how simulated policy performance tracks real-world outcomes. SimFoundry~\cite{ranawaka2026simfoundry} reconstructs scenes from video, supports rigid and articulated objects, and generates object, scene, and task variations called digital cousins. Its articulation module segments movable parts, generates URDF joints, and uses simulated motion and VLM feedback to refine them; it also estimates link mass, joint friction, and damping. GASE~\cite{zhang2026gase} and Mirage2Matter~\cite{gao2026mirage2matter} likewise combine Gaussian-splat appearance with interactive object geometry.

Other work addresses factors that determine simulation fidelity: joint refinement of geometry and calibration~\cite{moran2025splatting}, reconstruction for motion planning~\cite{sun2026highfidelity}, calibration and perturbation testing of a robot workspace~\cite{sedlacek2025realm}, and the effect of shadows and reflections on policy evaluation~\cite{zhu2026visually}. Genie Sim~3.0~\cite{yin2026genie} combines scanned environments with generated scenes, while MolmoB0T~\cite{deshpande2026molmob0t} studies transfer from large-scale procedural environments. SceneAgent focuses on coordinating reconstruction and physical modeling tools across input types and using agents to inspect and repair their outputs throughout conversion.

\begin{figure}[t]
\centering
\includegraphics[width=\linewidth]{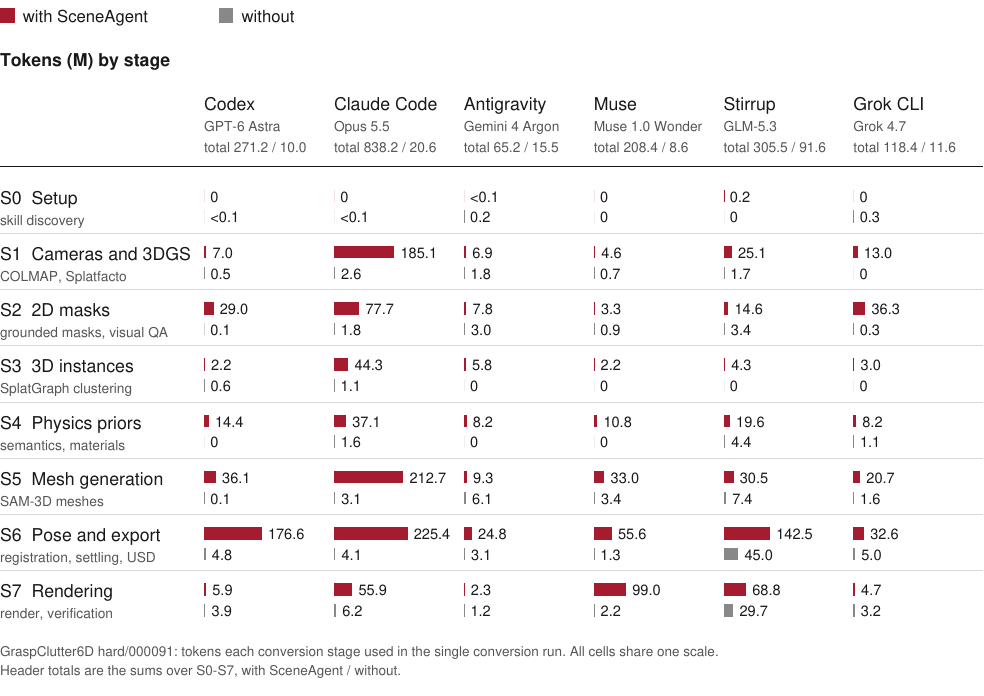}
\caption{Tokens used in each conversion stage of the GraspClutter6D bin scene (hard/000091), with and without SceneAgent skills.}
\label{fig:graspclutter-stage-tokens}
\end{figure}

\paragraph{Demonstration generation and scene variation.}
Digital reconstructions can expand limited demonstrations into larger training datasets. Real2Render2Real~\cite{yu2025real2render2real} synthesizes demonstrations from object scans and a human video using kinematic rendering. RoboSplat~\cite{yang2025demonstration} varies object pose, appearance, and embodiment while adapting demonstrated trajectories, and ExoGS~\cite{wang2026exogs} replays exoskeleton demonstrations through editable Gaussian-splat assets. ManiSplat~\cite{hu2026manisplat} reconstructs a scene from robot video to synthesize additional trajectories. TwinRL~\cite{xu2026twinrl} uses a reconstructed environment to expand fine-tuning data and identify configurations for human-in-the-loop reinforcement learning. Our demonstration workflow samples layouts, settles objects under physics, executes a scripted expert using simulator state, and retains successful episodes for policy fine-tuning.

\paragraph{Physical properties, articulation, and deformation.}
Visual captures leave many physical properties unobserved. Material-informed Gaussian Splatting~\cite{huynh2025materialinformed} transfers material labels to reconstructed geometry. Phys2Real~\cite{wang2025phys2real} combines VLM priors with interaction-based estimates, while D-REX~\cite{lou2026drex} identifies mass through differentiable simulation of observed interactions. ReconPhys~\cite{wang2026reconphys} estimates deformable dynamics from video, and 4DCodeBench~\cite{shen20264dcodebench} evaluates agents that reconstruct dynamic scenes from video as executable simulation and rendering programs, the setting of our 4D extension. Complementary methods address cable simulation~\cite{yang2026deformx}, articulated reconstruction~\cite{mishra2026artitwinsplat}, and geometry refinement using physical consistency~\cite{lee2026simuscene}. In this work, we do not aim to solve the inverse physics problem of accurately recovering physical parameters from input 3DGS. This problem can remain ill-posed even with robot--object interaction data. Instead, SceneAgent combines semantic features and VLM estimates to obtain approximate, physically plausible parameter values, intended as starting points for domain randomization of physical properties during policy training. We leave accurate physical parameter identification to separate future work.

\paragraph{Generative policy-evaluation environments.}
WorldGym~\cite{quevedo2025worldgym}, WorldEval~\cite{li2025worldeval}, and RoboWorld~\cite{jeon2026roboworld} investigate policy evaluation through generated video. SceneAgent constructs explicit geometry, physical parameters, and joints for execution in a physics simulator. This representation makes object properties available for inspection and editing.

\section{SceneAgent}
\label{sec:pipeline}

SceneAgent converts a capture into an interactive simulation environment through three stages: agent-guided reconstruction and repair, scene augmentation, and demonstration generation. We first describe the agent workflow, then the representations and conversion tools it coordinates, and finally the downstream training procedure. Section~\ref{sec:4d} extends the workflow from static 3D captures to videos of dynamic scenes.

\subsection{Agent Orchestration and Verification}
\label{sec:agents}

\begin{figure}[!htbp]
\centering
\makebox[\linewidth][c]{\includegraphics[width=\dimexpr\linewidth+2cm\relax]{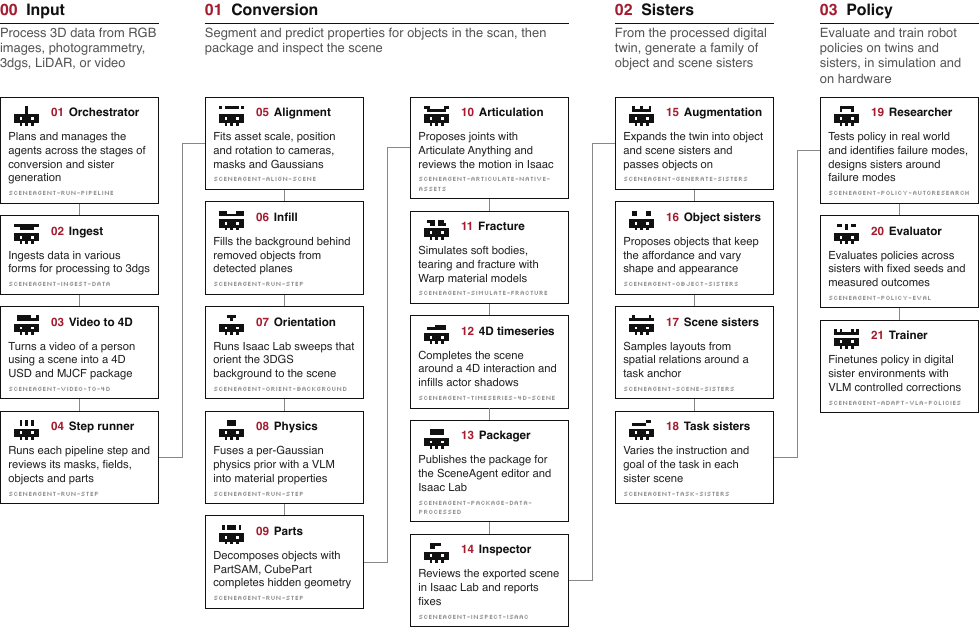}}
\caption{Agent workflow and available skills. The agents coordinate tools for capture processing, object reconstruction, physical modeling, and simulation, and inspect intermediate results to guide corrections.}
\label{fig:agents}
\end{figure}

Agents execute the conversion tools and inspect their outputs in the target simulator. This inspection is useful at transitions between models, where file conventions, coordinate frames, and segmentation errors can affect the assembled scene. For example, an object mesh may have the correct shape but appear at the wrong scale, orientation, or position after insertion into the environment. Agents render the scene, inspect these discrepancies, and correct the corresponding assets or transforms. Figure~\ref{fig:agents} summarizes the tools and skills available to the agents.

The orchestrator agent first plans the full conversion and processing steps from the input state, then the ingest agents prepare the input data and build the 3dgs, and finally the runner agent executes the processing tools with actor-critic visual review of each step. Alignment and orientation skills adjust object transforms and the background's placement in the simulator. A packager assembles the scene, and an inspector reviews the exported result in Isaac Lab. Object-, scene-, and task-variation skills then generate digital sisters and alternative layouts.

The workflow supports both scene-level conversion and inspection of individual objects. Agents can examine close-up renders to identify incomplete segmentation, missing geometry, or incorrect placement; Figure~\ref{fig:objects} shows examples of these detailed views. This iterative correction concerns the current conversion.

Each agent collaborates with the hardness and passes state through files in the scene directory. Following the orchestrator agent, they execute the conversion plan following a manifest for each step then a ledger of accepted or rejected outputs. Then per converted object in the scene, the reviewer agent inspects the object registration overlays, eight isolated views (six axis-aligned and two oblique), and finally two views of the object in the fully assembled scene. If a problem is discovered, such as an alignment, articulation, or decomposition issue, the agent retries by proposing a corrected pose or other fix and reviews again.

The alignment residuals in Figure~\ref{fig:agents} come from a trimmed similarity ICP, seeded with the SAM3D pose, that aligns each mesh to the scene. We use the normalized median and 90th-percentile point distances, coverage, and silhouette agreement, and accept a candidate only when all four pass fixed thresholds. Procedures from each conversion are saved to files that later conversions can read.

The agents also maintain SceneAgent's skills and tools and are instructed to research methods that may improve their conversion results, and if they identify a potential solution, request to evaluate it, and then modify their skills. The candidate suggested change to the tools and the current baseline are then run on a set of static 3D, dynamic 4D, and video-to-4D benchmarks, and the change is kept only if it clears fixed improvement and regression thresholds. For external users of a public release, a passing change becomes a draft pull request once the user approves it. This then initiates an independent benchmark run and maintainer review by the authors before it is merged.

\subsection{Input Preparation and Scene Representation}
\label{sec:representation}

For RGB image sequences, the pipeline uses COLMAP~\cite{schoenberger2016sfm} to estimate camera poses and Nerfstudio's \texttt{splatfacto} big~\cite{tancik2023nerfstudio,ye2024gsplat} to train a 3DGS reconstruction. Agents can inspect and correct capture-processing issues, including per-image color calibration~\cite{wang2024bilateral}. The pipeline also accepts existing 3DGS files, photogrammetry assets, and LiDAR captures, as well as generated scenes from models such as World Labs Marble~\cite{worldlabs2025marble}, NVIDIA Lyra~\cite{bahmani2025lyra}, and Tencent HY-World~\cite{hunyuanworld2025}. Videos of dynamic scenes follow the 4D pathway of Section~\ref{sec:4d}.

For the Gaussian-splat pathway, we represent the capture as $\{g_i\}_{i=1}^{N}$, with $g_i=(\boldsymbol{\mu}_i,\Sigma_i,\alpha_i,\mathbf{h}_i)$ denoting its center, covariance, opacity, and spherical-harmonic color coefficients. Calibrated cameras $c$ provide image projections $\pi_c$ and camera-frame depths $z_c$. The output is a scene $\mathcal{S}=(\mathcal{G}_{\mathrm{bg}},\{o_k\}_{k=1}^{K})$, containing a completed background splat and $K$ objects. Each object is represented as $o_k=(P_k,M_k,T_k,\theta_k,\mathcal{J}_k)$, where $P_k$ is its set of Gaussians, $M_k$ its mesh, $T_k$ its rotation, translation, and uniform scale, $\theta_k$ its physical properties, and $\mathcal{J}_k$ its joints, which may be empty.

\subsection{Object Reconstruction and Background Completion}
\label{sec:reconstruction}

We associate semantic features with the Gaussians using image masks and embeddings, following the general approach of LangSplat~\cite{qin2024langsplat}. SAM~\cite{kirillov2023sam} produces object masks on every tenth input view by default; across our scenes this stride ranges from 1 to 10. OpenCLIP ViT-B-16~\cite{radford2021clip,ilharco2021openclip} embeds each masked crop into a 512-dimensional vector, and DINOv2 ViT-S/14~\cite{oquab2023dinov2} gives each object a visual feature vector. Both are inputs to the physics prior (Section~\ref{sec:physics}); neither is used to decide which Gaussians belong to an object. Masks from different views are grouped into one object when they share 3D Gaussians: the centroids of their Gaussians lie within $\min(0.10,\,0.6\,e)$ reconstruction units of each other, where $e$ is the object's extent, and the two Gaussian sets have a containment of at least 0.18 or a Jaccard index of at least 0.08. SAM3D~\cite{al2025sam} generates object geometry.

To associate Gaussians with a visible object, we require each Gaussian center to project inside its mask and lie near the front surface. Specifically, Gaussian $i$ belongs to mask $m$ from camera $c$ when
\begin{equation}
\pi_c(\boldsymbol{\mu}_i)\in m
\quad\text{and}\quad
z_c(\boldsymbol{\mu}_i)\le \hat z_c(b_i)+\max\big(\epsilon_a,\ \epsilon_r\,\hat z_c(b_i)\big),
\label{eq:membership}
\end{equation}
where $\hat z_c(b_i)$ is the smallest camera-frame depth among Gaussian centers projecting into the same coarse image cell $b_i$, a square of $\mathrm{round}(s/192)$ pixels, at most 16, for an image whose shorter side is $s$ pixels. We use $\epsilon_a=0.08$ in reconstruction units and $\epsilon_r=0.03$. Because $\epsilon_a$ is in reconstruction units, its metric size depends on the capture's scale; in the kitchen, at 0.90--0.93~m per unit, it is about 7~cm. Let $A_m$ contain the Gaussians satisfying~\eqref{eq:membership}, and let $\mathcal{O}_k$ contain the masks grouped into object $k$. The object's Gaussians are then $P_k=\bigcup_{m\in\mathcal{O}_k}A_m$. An optional camera-consensus mode retains only Gaussians assigned from at least $\min(\kappa,C_k)$ of the $C_k$ cameras observing the object, for a chosen threshold $\kappa$. Our experiments use $\kappa=1$, so $P_k$ is the full union. Each generated mesh is aligned to its object's Gaussians $P_k$ by a trimmed similarity ICP with a single uniform scale, initialized from the SAM3D pose.

After removing the foreground Gaussians, we complete exposed background regions using detected planes. New Gaussians receive colors from nearby plane surfaces in the input images, with optional LaMa texture inpainting~\cite{suvorov2022lama}. The generated object meshes are assembled with the completed background and inspected in the simulator.

\subsection{Physical Properties, Parts, and Articulation}
\label{sec:physics}

For each object, we combine semantic features, predictions from a per-Gaussian physics model, and a vision-language model (VLM) to estimate material and physical properties. These include friction, rigidity, mass, and density. The implementation uses 10 material classes, 6 body types, and 8 collision primitives, and records the resulting properties in the exported scene.

The physics prior's per-Gaussian class logits and property estimates are averaged over $P_k$. A provisional material class is selected by a weighted vote,
\begin{equation}
y_k^\star=\arg\max_{y}\sum_{v\in V_k} w_v\,\mathbf{1}[y_v=y],
\label{eq:material-vote}
\end{equation}
where each source $v\in V_k$ proposes class $y_v$ with weight $w_v$. Sources comprise the physics prior, the VLM, and keyword priors derived from the object name. Weights use source confidence or a fixed value for keyword priors and are scaled by 0.4 when $y_v$ is \emph{unknown}. A subsequent VLM audit and recorded material review may replace the provisional class.

The physics prior is a multilayer perceptron whose inputs are DINOv2 and OpenCLIP features and 16 geometric quantities. It has output heads for material class, body type, and collision primitive, and it regresses nine numerical properties. It is trained on object records from GSO, YCB, and ABO, and its targets are labels written by a VLM, not physical measurements. The VLM audit decides the material, and we treat the numerical values as approximate priors.

The model regresses mass and density as logarithms. Young's modulus, Poisson's ratio, and the rendering material parameters are looked up from the material class. Rigidity is a unitless score between 0 and 1. It enters the simulation only through deformability, and an object becomes a soft body when its deformability is at least 0.65.

The ten material classes are plastic, metal, wood, fabric, paper and cardboard, rubber and foam, glass and ceramic, food and organic matter, liquid and granular matter, and unknown. The six body types are dynamic rigid, static rigid, kinematic, soft deformable, cloth shell, and fluid or granular. The eight collision primitives are box, sphere, cylinder, capsule, convex hull, triangle mesh, tetrahedral mesh, and cloth mesh. The Isaac Lab export uses mass, friction, restitution, body type, and collision approximation, and it stores the remaining properties as USD metadata. The recorded material review is automatic. It is signed by a coding agent, not a person.

\begin{figure}[!htbp]
\centering
\includegraphics[width=\linewidth]{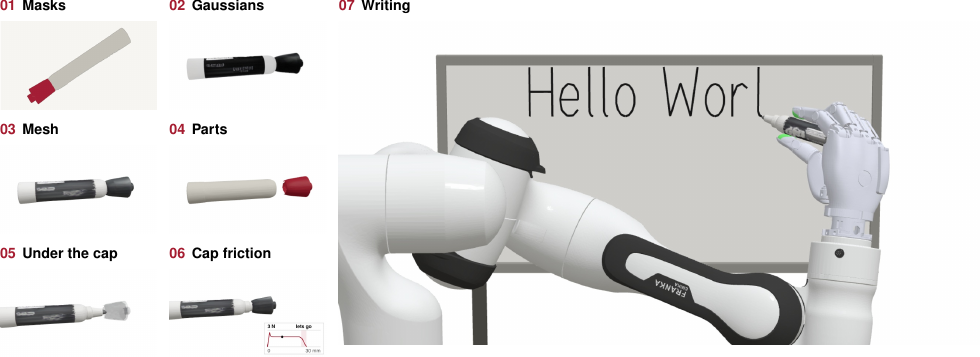}
\caption{Marker reconstruction and interaction. A single product image is converted into object geometry and parts, including completion of the tip beneath the cap. Two Franka Emika Panda arms with Sharpa Wave hands manipulate the resulting marker and write on a whiteboard in simulation.}
\label{fig:marker}
\end{figure}

PartSAM~\cite{zhu2025partsam} decomposes each reconstructed object into parts. A VLM inspects these parts for missing geometry, such as the tip of a marker hidden beneath its cap. When completion is needed, CubePart~\cite{zhu2026cubepart} generates the missing region, which is appended to the mesh. Articulate Anything~\cite{le2025articulateanything} then proposes joints and motion ranges in URDF format from image or text inputs. Agents review the resulting articulation in simulation. Figure~\ref{fig:marker} follows this process from a marker image to a simulated writing interaction.

\begin{figure}[!htbp]
\centering
\includegraphics[width=\linewidth]{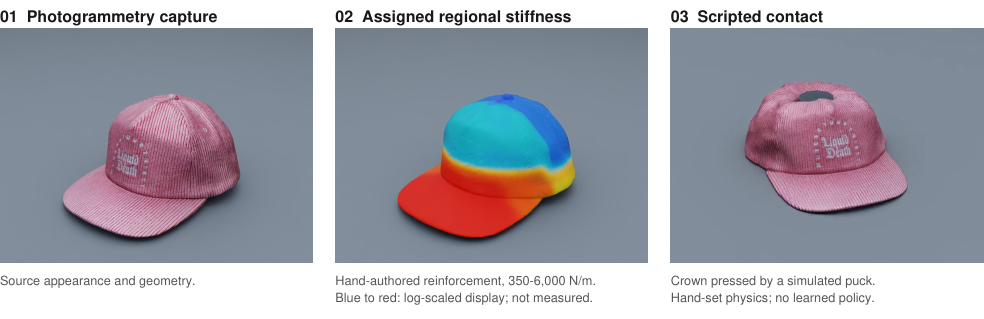}
\caption{A corduroy cap captured with photogrammetry, annotated with estimated rigidity, and simulated under contact. The example illustrates the conversion from visual geometry to a deformable asset.}
\label{fig:hat}
\end{figure}

\begin{figure}[!htbp]
\centering
\includegraphics[width=\linewidth]{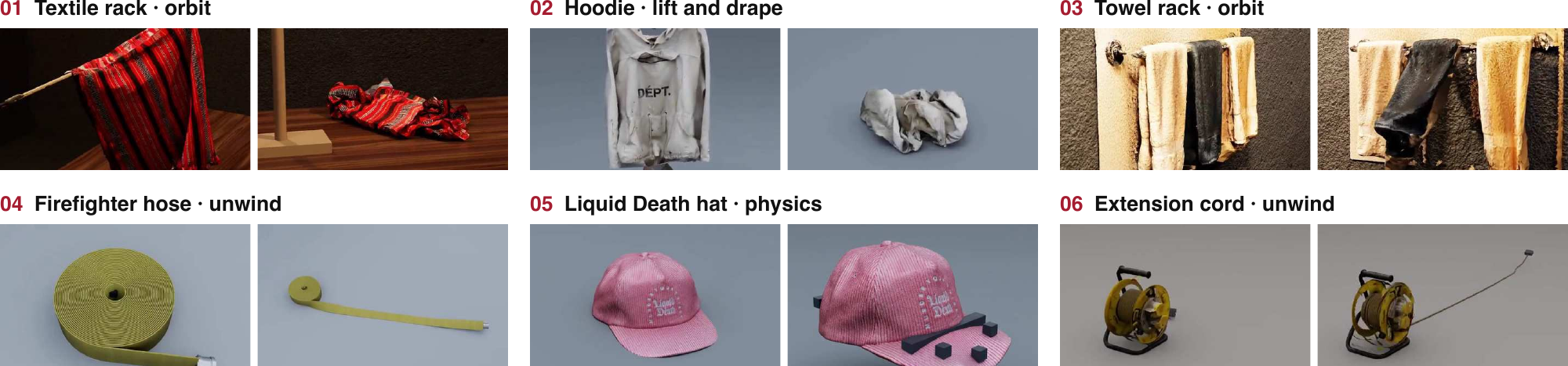}
\caption{Deformable assets converted from Polycam~\cite{polycam} captures. Each object is shown at rest and under simulated deformation.}
\label{fig:deformables}
\end{figure}

Objects requiring deformation are represented as soft bodies. Figure~\ref{fig:hat} illustrates the estimated rigidity and simulated deformation of a corduroy cap, and Figure~\ref{fig:deformables} shows additional objects at rest and under deformation.

\begin{figure}[!htbp]
\centering
\includegraphics[width=\linewidth]{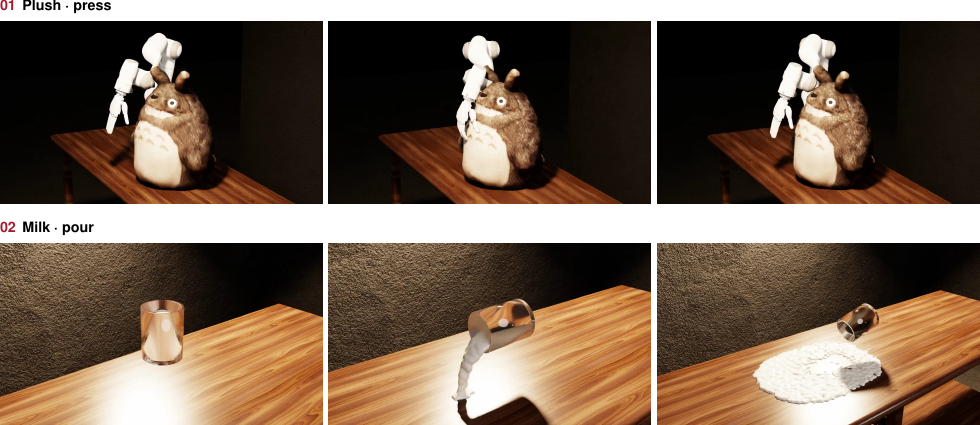}
\caption{Contact and fluid simulation. A Franka Emika Panda arm with a Sharpa Wave hand presses a plush toy (01). A cup tips and pours milk onto a table (02). The milk is simulated with MPM.}
\label{fig:soft-fluid}
\end{figure}

\subsection{Fracture Simulation}
\label{sec:destruction}

\begin{figure}[!htbp]
\centering
\includegraphics[width=\linewidth]{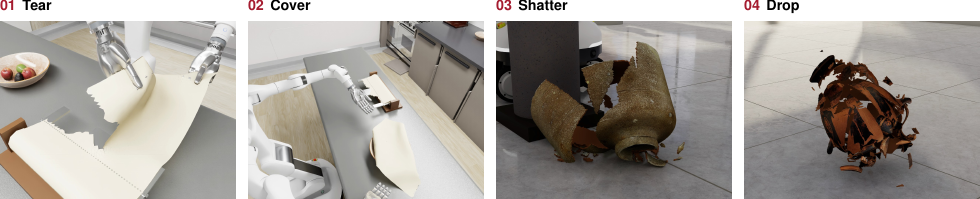}
\caption{Fracture simulation. A Dexmate Vega with two Sharpa Wave hands tears a parchment sheet from a roll and covers a bowl of fruit (01--02). Ceramic jars fracture after being knocked from a pedestal or dropped (03--04). The parchment parameters are engineering estimates.}
\label{fig:destruction}
\end{figure}

We also explore material failure during manipulation, including paper tearing and ceramic fracture. Figure~\ref{fig:destruction} shows a Dexmate Vega with two Sharpa Wave hands tearing parchment from a roll and placing it over a bowl, alongside examples of jars shattering after impact. The parchment uses an orthotropic thin-sheet finite-element model in NVIDIA Warp~\cite{macklin2022warp}, with irreversible cohesive fracture along interior edges. The tear path emerges as interfaces fail. In the illustrated rollout, 49 interfaces fail progressively, leaving a detached sheet that covers 95\% of the bowl opening after the hands withdraw. The strength, fracture energy, and friction parameters are currently engineering estimates; prediction from captures and validation against measured break forces remain future work.

Coverage is the fraction of points on a 71 by 71 grid over the bowl opening that lie beneath the detached sheet, counting a point only where the sheet is no lower than 12~mm below the rim, and we report the lowest value over the frames after the hands withdraw. The ceramic examples use a separate model. Each scanned vessel is divided into connected patches of its original surface triangles, an inner wall 4.5~mm thick is inferred for each patch, and neighboring patches are joined by bonds in NVIDIA Blast that break irreversibly once the stress in the wall exceeds the material strength. Cracks therefore follow the boundaries between patches, and the fragments are native PhysX rigid bodies.

The material values are engineering assumptions that a scan cannot measure, with a tensile strength of 6~MPa, a Young's modulus of 6~GPa, a fracture energy of 14~J/m$^2$, and a density of 1850~kg/m$^3$ for every vessel. Neither model is part of the automatic scene export. Each example is authored as its own Isaac Lab scene, the parchment in the kitchen scene and the vessels from their Polycam captures, which a preparation script converts to USD and divides into fragments. All examples use scripted controllers, and none uses teleoperation or a learned policy. The robot hands tear and place the parchment, a hand holding a cloth wipes one vessel and pushes it off its pedestal, and the other vessels are dropped without a robot.


\begin{figure}[!htbp]
\centering
\includegraphics[width=\linewidth]{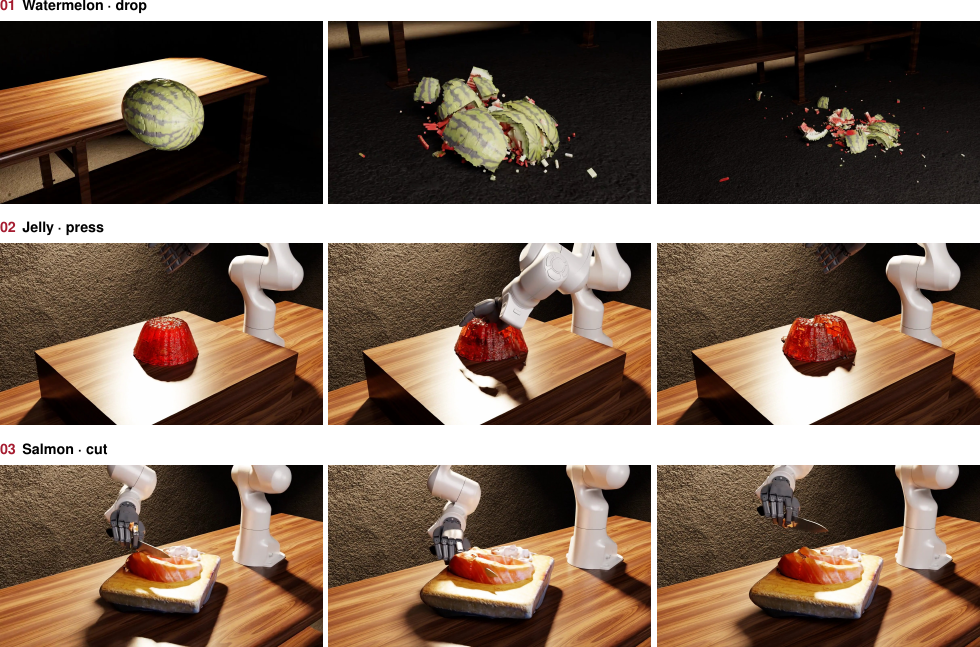}
\caption{Fracture and cutting. A watermelon breaks apart on impact after a drop (01). A Franka Emika Panda arm with a Sharpa Wave hand presses into a jelly, which cracks (02), and cuts salmon with a chef's knife (03). The watermelon and jelly use cohesive finite elements in Newton.}
\label{fig:fracture-cutting}
\end{figure}

\subsection{From 3D to 4D Captures}
\label{sec:4d}

The pipeline described so far converts static captures. A video of a dynamic scene also shows how the scene moves, such as liquid flowing around obstacles, a gel deforming under impact, or a fruit breaking apart. We refer to such a video as a 4D capture. SceneAgent converts 4D captures into simulations that reproduce the observed motion. The agents follow the same workflow of planning, tool execution, rendering, and repair as for 3D captures (Section~\ref{sec:agents}), while the scene representation, the estimation of physical parameters, and the inspection of results are extended to sequences of frames.

The input is a sequence of frames $\{I_t\}_{t=1}^{F}$ from a camera with unknown intrinsics $K$ and extrinsics $E$, which are estimated from the video. We extend the scene of Section~\ref{sec:representation} to $\mathcal{S}=(K,E,\mathcal{G},\{d_m\})$, where $\mathcal{G}$ is the static geometry, such as containers, obstacles, and the surrounding room. Each dynamic material $d_m=(\mathbf{x}_m^{0},\theta_m,\Phi_m)$ has an initial state $\mathbf{x}_m^{0}$, physical parameters $\theta_m$, and a simulator $\Phi_m$ that advances its state from one frame to the next,
\begin{equation}
\mathbf{x}_m^{t+1}=\Phi_m\big(\mathbf{x}_m^{t},\theta_m,\mathcal{G},\{\mathbf{x}_n^{t}\}_{n\neq m}\big),
\label{eq:dynamics}
\end{equation}
with contact between the material, the static geometry, and the other materials. At every frame, the scene exports a triangle mesh $M_m^t$ for each material together with the positions of its particles or vertices. Each particle or vertex keeps its index over the whole sequence, which gives persistent trajectories $X_m\in\mathbb{R}^{F\times N_m\times 3}$. Rendering the scene with the estimated camera produces frames $\hat I_t$ that are compared with the input frames $I_t$.

SceneAgent's 4D skills cover video-to-4D conversion, 4D scene completion, and fracture simulation. The agent first extracts reference frames from the video and estimates the camera, for example from the vanishing points of a container's edges and from object silhouettes, and checks the estimate by projecting the static geometry onto the frames. It then builds the static geometry and selects a simulator for each material. Gels, elastoplastic solids, and liquids interacting with rigid bodies are simulated with the material point method (MPM)~\cite{sulsky1994particle,hu2018mlsmpm}, incompressible liquids with FLIP~\cite{zhu2005animating}, and brittle objects with fracture models. The agent fits the initial states and physical parameters to the motion in the video. Liquid and deformable surfaces are extracted from the simulated particles at every frame~\cite{zhu2005animating}. Particles are not deleted or reseeded during the simulation, so each trajectory follows the same piece of material for the whole video.

To check its results, the agent renders the simulated sequence with the estimated camera and compares it with the input video in side-by-side and overlay views at early, middle, and late frames. Depending on the discrepancy, it revises the camera, the static geometry, the initial states, or the physical parameters and compares the views again. The final scene is produced by a single build script that does not read the input video or any external assets, so its frames, meshes, and trajectories can be regenerated from scratch and checked for format, frame count, and resolution.

Unlike the priors used for static captures (Section~\ref{sec:physics}), the initial states and physical parameters of a 4D capture are constrained by the observed motion, and the agent adjusts them until the simulation matches the video. The fitted parameters reproduce the observed motion, but they are not necessarily unique and have not been tested on other interactions. The solvers are of the same types as those in our deformable, fluid, and fracture examples (Figures~\ref{fig:soft-fluid}--\ref{fig:fracture-cutting}).

\subsection{Scene Export and Editing}
\label{sec:export}

\begin{figure}[!htbp]
\centering
\includegraphics[width=\linewidth]{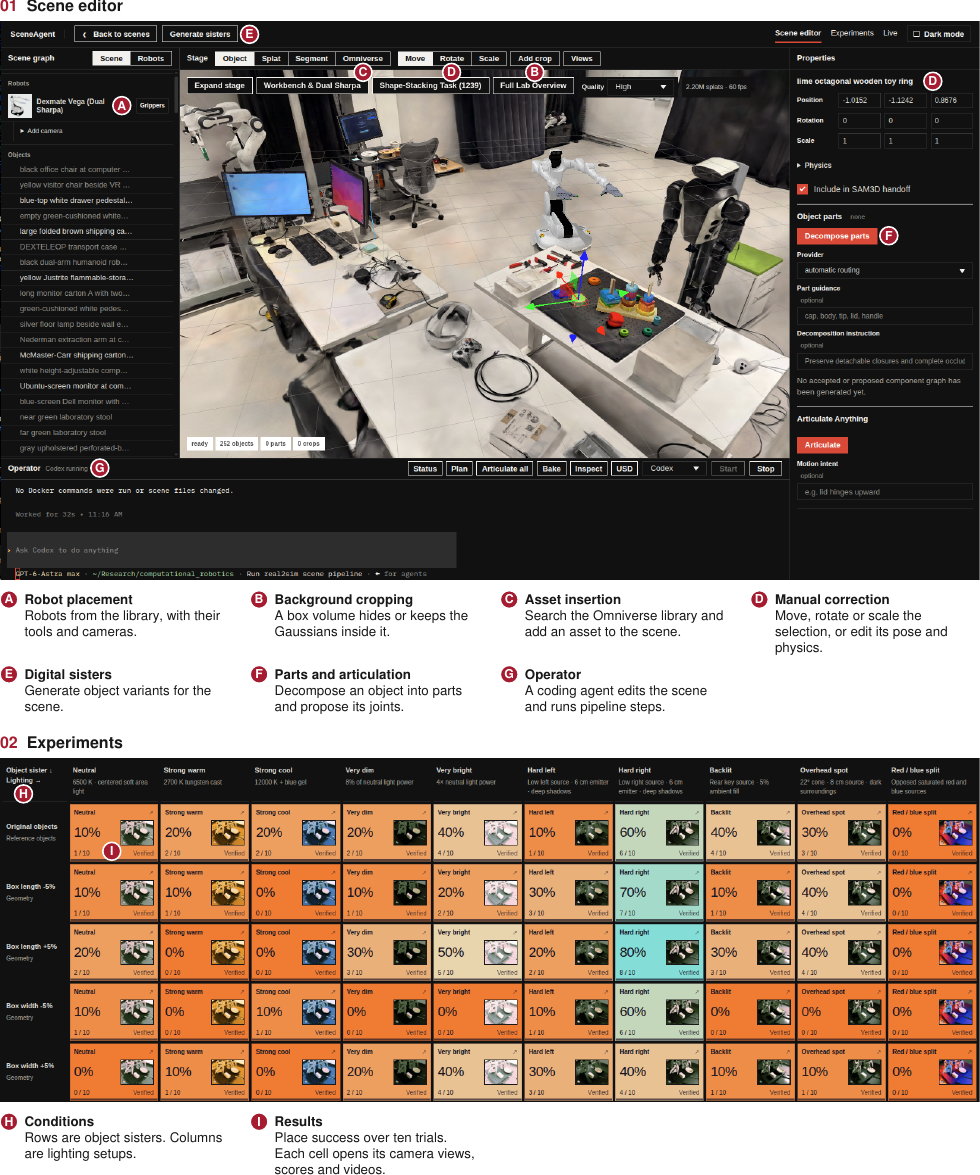}
\caption{The interactive viewer. (01) The scene editor for the lab scene with a Dexmate Vega, with the scene graph, the Three.js stage, the properties of the selected object, and the agent operator. Lettered callouts mark its main operations. (02) The experiment view of a digital-sister evaluation. Each cell is one object sister under one lighting setup and reports place success over ten trials.}
\label{fig:scope}
\end{figure}

The converted scene is packaged in Universal Scene Description (USD) format~\cite{pixar2016openusd}. Our experiments use Isaac Lab~\cite{mittal2025isaaclab}; the pipeline also targets MuJoCo~\cite{todorov2012mujoco} and Unreal Engine~\cite{epic2024unreal}. An interactive Three.js viewer (Figure~\ref{fig:scope}) supports robot placement, background cropping, asset insertion, and manual correction. Users can add converted objects or assets from repositories such as NVIDIA Omniverse~\cite{nvidia2024omniverse} before using the scene for a task.

SceneAgent exports each scene as OpenUSD. We worked primarily in Isaac Lab, and we also tested and verified the exported scenes in MuJoCo and Unreal Engine. OpenUSD's physics schema describes rigid bodies, mass, collision shapes, joints, and physics materials with static and dynamic friction, restitution, and density, so a simulator that reads the schema can load these properties from the file. Our export marks each movable object as a rigid body with its mass and collision geometry, and stores its friction, restitution, density, and other estimated properties as attributes on the same object.

The schema does not cover deformable bodies, fracture, or fluids and gels. Isaac Lab adds these through its PhysX extensions and through NVIDIA Warp, Newton, and Blast, which our soft-body, tearing, fracture, and fluid examples use. MuJoCo and Unreal Engine have their own deformable and destruction models, so these behaviors do not carry over through the USD file and would need to be set up in each engine.

\subsection{Digital Twins and Sisters}
\label{sec:sisters}

\begin{figure}[!htbp]
\centering
\includegraphics[width=\linewidth]{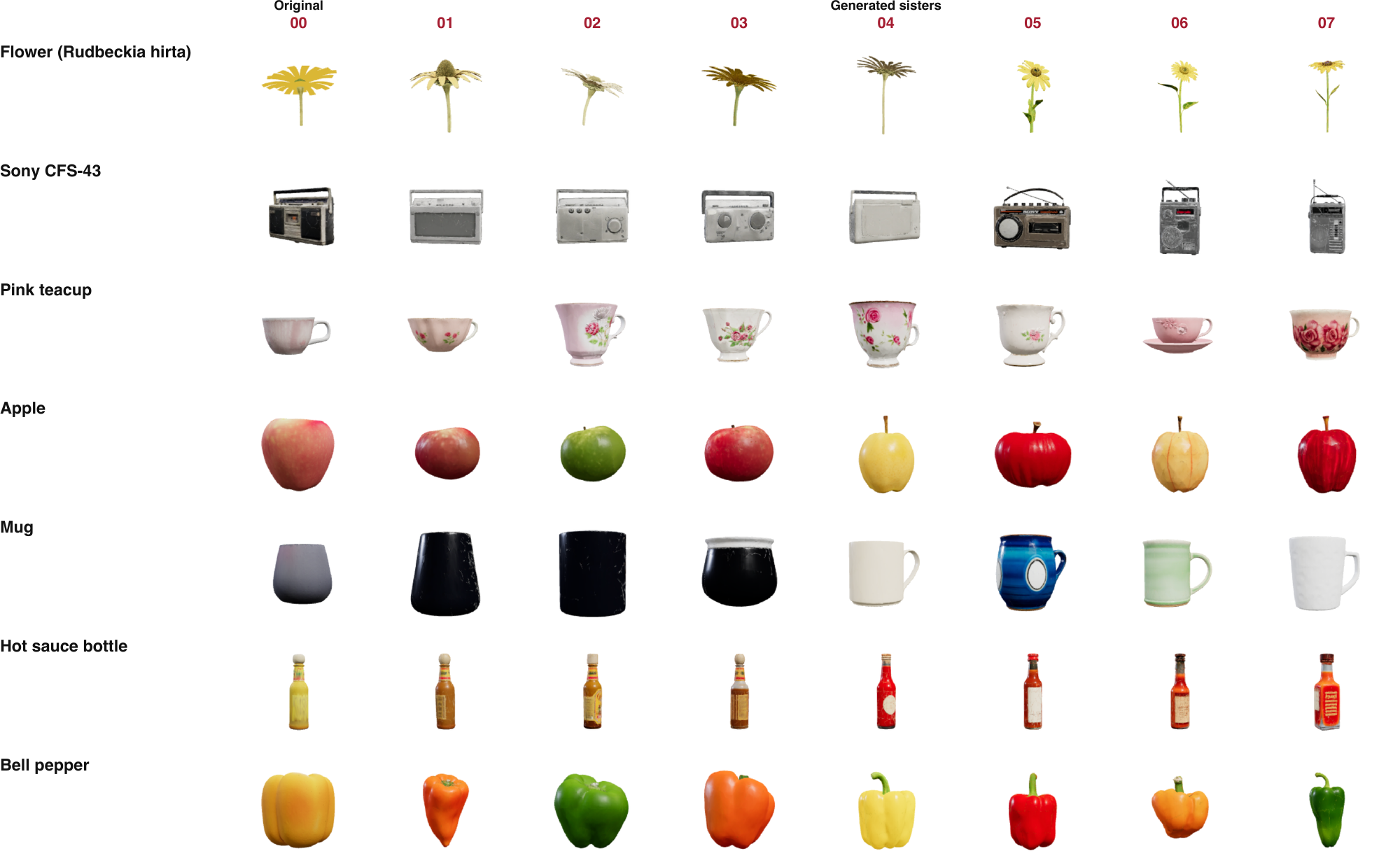}
\caption{Digital sisters of reconstructed objects. Each row shows one converted object followed by seven generated variants.}
\label{fig:sisters}
\end{figure}

The reconstructed scene provides a digital twin of the captured workspace. To introduce variation, we generate related instances of its objects, termed \emph{digital sisters}, and compose scenes with randomized object placement and lighting. These variations aim to expose a policy to a broader range of appearances and configurations during training. Figure~\ref{fig:sisters} shows seven reconstructed objects and their generated variants.

We generate digital sisters in two ways. The first perturbs the twin itself without calling a model. Bounded deformations change the shape of the object's own mesh, and each result is checked for changes in extent and volume and for whether the object can still be grasped. A procedural texture of noise, hue rotation, mottling, and speckle is baked into the vertex colors. The second uses Meshy. For new object instances, a reasoning model (Gemini 2.5 Flash) divides the object into functional components and proposes variations of each, and an image model (Gemini 3 Pro Image) edits the object image one component at a time. Each edit is then checked for real-world plausibility and fit with the scene, and Meshy's image-to-3D model lifts each accepted image to a textured mesh. For appearance alone, Meshy's retexture model (Meshy-6) repaints the original mesh with its geometry and UVs unchanged.

The edit prompt asks the image model to change one named component by the proposed variation, keep the result realistic and ordinary, leave the other components unchanged, and match the scene. The retexture prompt asks for the same object with a small change in surface detail, such as fine grain or light wear, and rules out new components, holes, and changes in shape.

\subsection{Hybrid Scenes from Captures and Digital Art}
\label{sec:hybrid}

SceneAgent can also convert an individual captured object and place it in an artist-authored environment. This allows a task-specific asset, such as a device component or server-rack part, to be incorporated into an existing scene without recapturing the entire workspace. The same object conversion, physical modeling, and inspection tools apply to these hybrid environments.

\begin{figure}[!htbp]
\centering
\includegraphics[width=\linewidth]{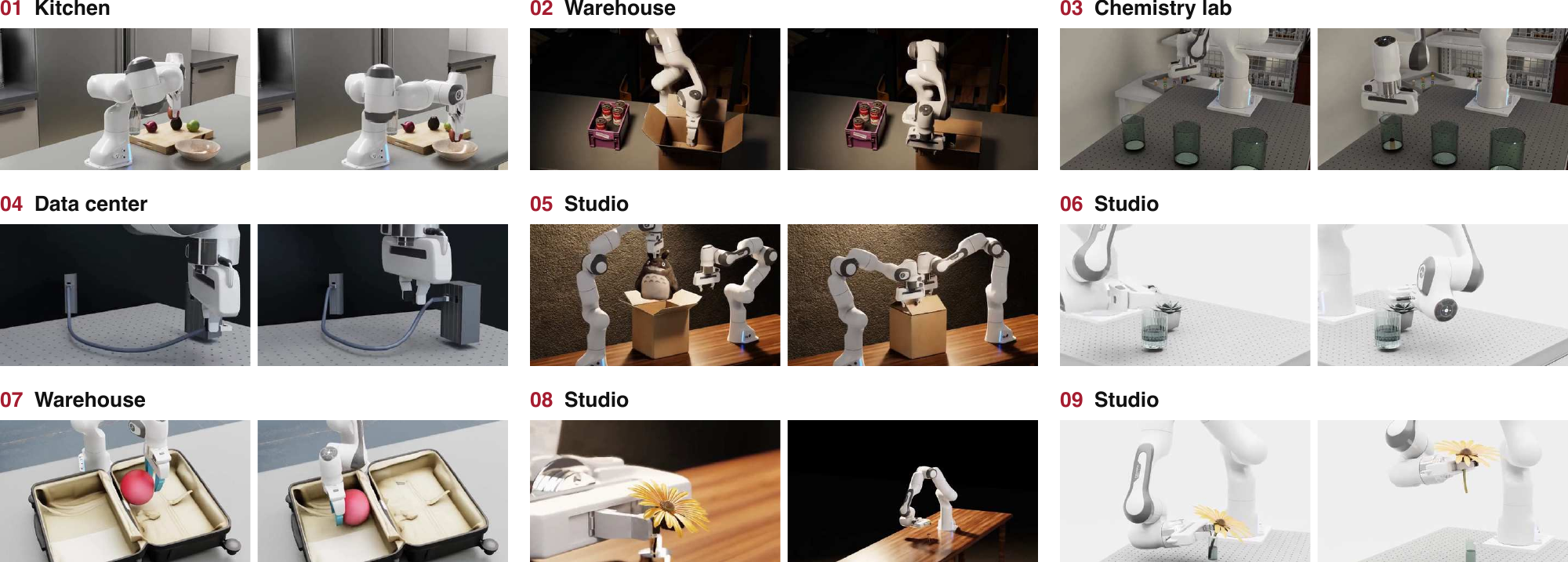}
\caption{Hybrid environments combining captured objects with artist-authored assets. Franka Emika Panda arms with UMI grippers are shown during and after contact-rich and deformable-object tasks in simulation.}
\label{fig:umi}
\end{figure}

\begin{figure}[!htbp]
\centering
\includegraphics[width=\linewidth]{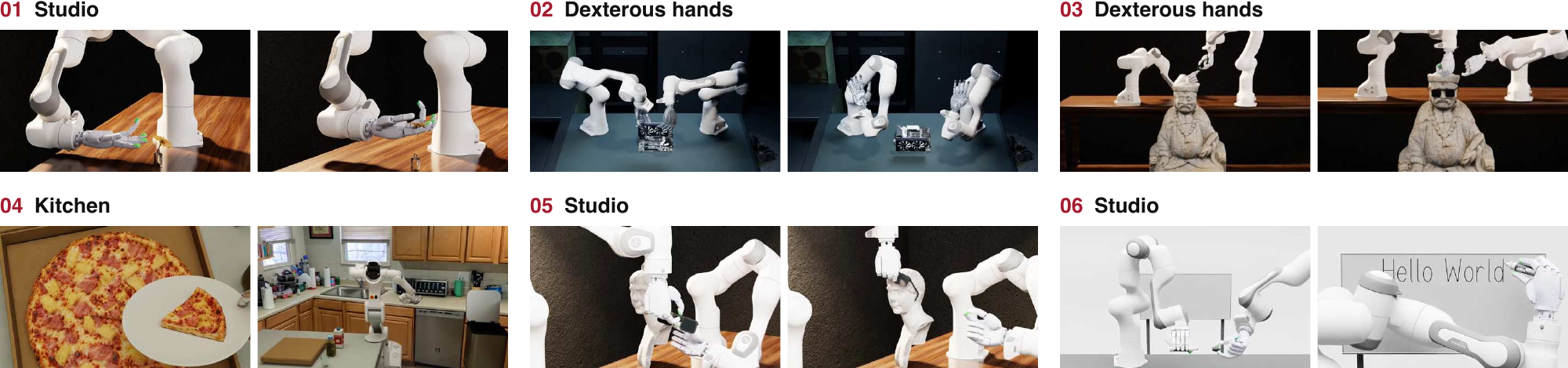}
\caption{Dexterous manipulation in converted environments. Sharpa hands mounted on Panda arms and a Dexmate mobile manipulator interact with reconstructed objects during multi-step simulated tasks.}
\label{fig:sharpa}
\end{figure}

The examples span grippers and dexterous hands. Figure~\ref{fig:umi} shows Franka Emika Panda arms with Universal Manipulation Interface (UMI) grippers~\cite{chi2024umi}, while Figure~\ref{fig:sharpa} shows Sharpa hands on Panda arms and a Dexmate mobile manipulator. These examples illustrate the range of environments and interactions that can be assembled; the policy-transfer study in Section~\ref{sec:transfer} focuses on pick-and-place tasks.

\begin{figure}[!htbp]
\centering
\includegraphics[width=\linewidth]{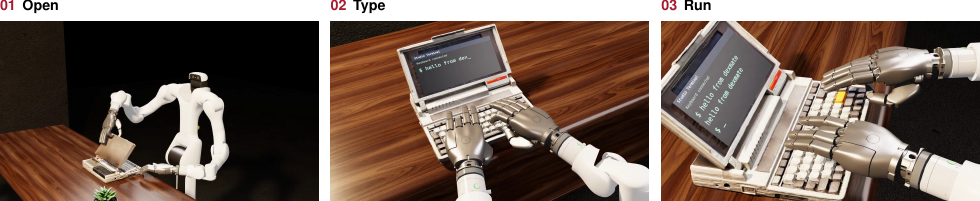}
\caption{A Dexmate Vega opens a laptop (01) and types a command on its keyboard (02--03).}
\label{fig:dexmate-laptop}
\end{figure}

\subsection{Demonstration Generation and Policy Learning}
\label{sec:training}

Given a converted workspace and a robot configuration, the training workflow generates scripted expert demonstrations and uses them to fine-tune a pretrained vision-language-action (VLA) policy. The demonstration generator uses ground-truth simulator state to plan actions and renders observations from the robot's cameras.

\begin{figure}[!t]
\begingroup
\raggedright
{\large\bfseries Agentic Policy Evaluation and Training\par}
\vspace{2pt}
{\small\color{black!65}Automate evaluations and repair failure modes in digital sisters of Tasks and Scenes\par}
\endgroup
\vspace{4pt}
\centering
\includegraphics[width=\linewidth]{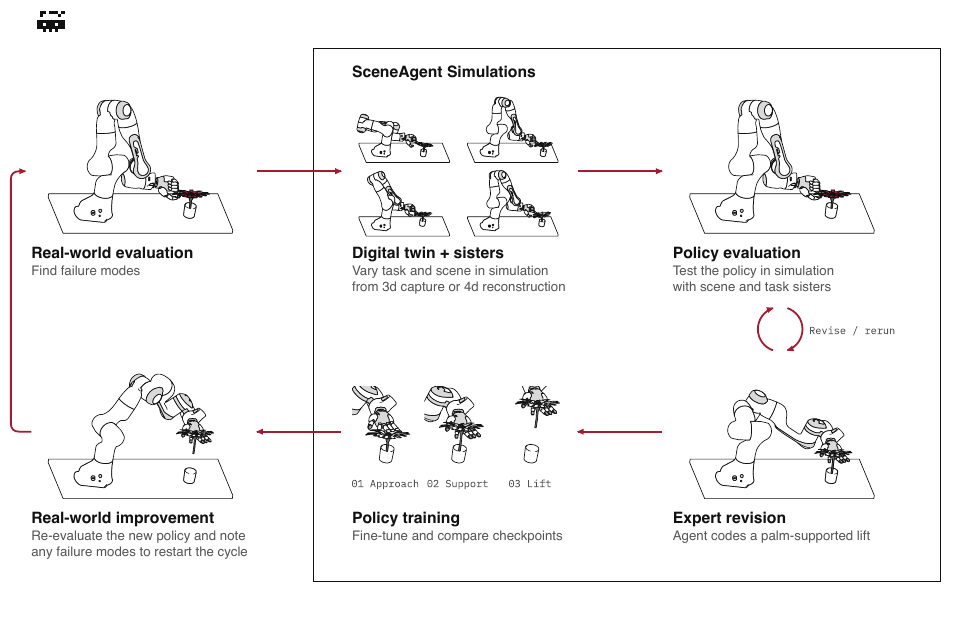}
\caption{Agentic real2sim2real policy autoresearch, illustrated with a Franka Emika
Panda and Sharpa Wave hand. First, real task recordings expose failures then agents construct
a digital twin, generate task and scene sisters, and evaluate the current policy.
Coding agents revise expert controllers and rerun simulated tasks. Task-verified
recordings feed policy fine-tuning, and fixed paired validation trials compare
checkpoints before real-world re-evaluation.}
\label{fig:autoresearch}
\end{figure}

\paragraph{Policy autoresearch loop.}
Policy training and evaluation optionally has agents that run in an automated autoresearch loop including identifying failure modes in reality or simulation, creating simulation digital sister task or scene simulations, and then correcting failures with a VLM to steer robot performance in simulation (Figure~\ref{fig:autoresearch}). The current policy is first rolled out on the real robot or in simulation, and agents review the recordings to find its failure modes and mark the segments where they occur. For each failure mode, an agent proposes likely causes and designs digital sisters that reproduce them, varying the relevant objects, placement, or lighting. The policy is then evaluated in these sisters. Where it fails, a coding model (GPT-6 Astra or Gemini 4 Argon) writes or revises the robot controller that serves as the expert for those cases. The controller is run in the sisters, and only episodes that pass the task's success check are kept. These simulated recordings are added to the fine-tuning data, the policy is retrained, and the new checkpoint is compared with the previous one on a fixed set of paired validation trials before it returns to the real robot for the next round.
\paragraph{Initial-state randomization.}
We vary object placement and size and perturb the camera and robot-base poses around their calibrated configurations. Objects are sampled within the task region and allowed to settle under simulated physics before execution. This provides physically settled initial configurations for demonstration collection.

\paragraph{Scripted expert and episode filtering.}
The expert adapts a pick-and-place routine to the sampled object poses: approach, descend, grasp, lift, carry, lower, release, and retreat. A success check determines whether the object ends inside the target container, and unsuccessful episodes are discarded. Observations are rendered by depth-compositing the mesh foreground, including the robot and manipulated objects, over the Gaussian-splat background. In the loop, later versions of the expert are the controllers written by the coding model, and they pass the same success check.

For each platform and task across the Franka Panda, the dexterous hands, and the Dexmate mobile manipulator, the controller is written by a VLM (GPT-6 Astra or Gemini 4 Argon), and the same approach is used in simulation and on the real robot due to the well-documented spatial understanding capabilities of these models. Platform-specific behavior, such as finger coordination in dexterous grasps and the coordination of the mobile base with the arms, is part of each generated controller.

\paragraph{Policy fine-tuning.}
Camera observations are paired with expert actions and a natural-language instruction sampled from alternative task descriptions. The resulting demonstrations are used for LoRA fine-tuning~\cite{hu2022lora} of a pretrained VLA. The reported fine-tuning data are generated in simulation; the starting policy retains its pretraining history. The real-world deployment setup and current evaluation scope are described in Section~\ref{sec:transfer}.



\FloatBarrier
\section{Results}
\label{sec:evaluation}

We first evaluate reconstruction quality on both a shared kitchen capture and a cluttered bin scene, then on dynamic scenes from 4DCodeBench, and finally present qualitative conversion examples and preliminary real-world policy transfer.

\subsection{Reconstruction Quality Across Agent Harnesses}
\label{sec:harnesses}

\paragraph{Setup and reference.}
Each harness converted the same scene from the MessyKitchens benchmark~\cite{ansari2026messykitchens}, once with SceneAgent skills and once without them. We score the delivered objects against the benchmark's ground-truth meshes, which come from 3D scans of the real objects and are independent of every conversion run, including ours. We use these meshes as the benchmark provides them, without postprocessing or manual correction. Figure~\ref{fig:harness} lists the model and reasoning setting of each harness. Every run receives the same task prompt and the same version of the SceneAgent skills, and no person intervened during the runs.

Each delivered mesh is placed using its saved transform. A single rigid transform derived from a shared capture frame aligns the run to the reference; object scales are preserved, and no object is individually refit or repaired for scoring. We sample 100,000 points per object with three sampling seeds, average metrics across seeds, and weight objects equally. These seeds vary metric sampling, not the scene-conversion run.

We also evaluate on scene hard/000091 of GraspClutter6D~\cite{back2025graspclutter6d}, in which 20 YCB objects are stacked and occlude one another in a bin. The input is 13 RGB views at 1920$\times$1080. We score the delivered objects against the dataset's ground-truth object meshes, placed with its registered 6D poses. Each harness again converted the scene once with SceneAgent skills and once without them. Both conditions receive the same task prompt, except that the with-skills prompt also instructs the agent to use the SceneAgent skills. We also run the official SimFoundry pipeline~\cite{ranawaka2026simfoundry} on the same 13 images.

Each run is aligned to the reference with a single similarity transform, and no object is individually refit or repaired for scoring. We sample 10,000 surface points per ground-truth object and match delivered objects to ground-truth objects with the Hungarian algorithm, using Chamfer distance as the matching cost.

\paragraph{Metrics.}
For each object and seed, let $X$ be points sampled from the delivered mesh, $Y$ the fixed reference points, and $d(x,Y)=\min_{y\in Y}\|x-y\|$. We report the symmetric mean nearest-point distance,
\begin{equation}
d_{\mathrm{CD}}(X,Y)=\frac{1}{2}\Big(\frac{1}{|X|}\sum_{x\in X} d(x,Y)+\frac{1}{|Y|}\sum_{y\in Y} d(y,X)\Big),
\label{eq:chamfer}
\end{equation}
and F1 at threshold $\tau$,
\begin{equation}
F_\tau=\frac{2P_\tau R_\tau}{P_\tau+R_\tau},\quad
P_\tau=\frac{1}{|X|}\sum_{x\in X}\mathbf{1}[d(x,Y)\le\tau],\quad
R_\tau=\frac{1}{|Y|}\sum_{y\in Y}\mathbf{1}[d(y,X)\le\tau],
\label{eq:f1}
\end{equation}
with $\tau=1$~cm and $F_\tau=0$ when $P_\tau+R_\tau=0$. On the kitchen capture, both metrics are averaged over sampling seeds and then over objects. On GraspClutter6D, F1 is averaged over all 20 ground-truth objects, with unmatched objects scoring zero, and Chamfer distance is averaged over matched objects.

\paragraph{Paired comparisons.}
On the kitchen capture, SceneAgent improves both metrics for every harness with scorable outputs in both conditions. F1 at 1~cm increases from 16.45\% to 41.13\% for Codex, from 7.07\% to 19.29\% for Claude Code, and from 6.33\% to 17.97\% for ZCode. Their Chamfer distances decrease by 1.35--2.61~cm. Grok Build with SceneAgent produces three of the four objects, omitting the knife block. On the matched three-object subset, its F1 increases from 5.67\% to 17.68\%, and its Chamfer distance decreases from 5.25 to 2.81~cm.

On GraspClutter6D, SceneAgent improves both metrics for every harness (Figure~\ref{fig:harness}). F1 at 1~cm increases from 9.51\% to 60.75\% for Codex, from 5.58\% to 60.46\% for Claude Code, from 2.36\% to 52.25\% for Antigravity, from 7.19\% to 52.70\% for Grok CLI, from 5.75\% to 41.39\% for Muse, and from 6.93\% to 40.45\% for GLM-5.3. Chamfer distance decreases from 6.20--9.89~cm without SceneAgent to 1.43--3.43~cm with it. SimFoundry reaches 6.9\% F1 and a 4.29~cm Chamfer distance, worse than every SceneAgent configuration on both metrics.

\paragraph{Incomplete runs and baseline comparison.}
Without SceneAgent, Antigravity and Muse produce no object geometry that can be scored. With SceneAgent, their F1 scores are 20.58\% and 5.14\%, respectively. SimFoundry~\cite{ranawaka2026simfoundry} achieves 10.21\% F1 and an 11.67~cm Chamfer distance on the same capture. Several SceneAgent-assisted configurations therefore exceed its F1 score under this protocol, although Muse does not. These comparisons concern reconstruction geometry in this kitchen, with respect to the stated reference.

\subsection{Dynamic Scene Reconstruction on 4DCodeBench}
\label{sec:4dcodebench}

\paragraph{Setup.}
4DCodeBench~\cite{shen20264dcodebench} evaluates agents that reconstruct dynamic scenes from video as executable simulation and rendering programs. We use three of its synthetic cases, whose reference worlds provide 3D ground truth. They are a liquid dam break around rigid cylindrical pillars in a transparent tank (B\_09, 130 frames at $1280\times720$), a block of gel or liquid and a rigid sphere interacting in a transparent tank (Z\_06, 150 frames at $1280\times720$), and a layered watermelon breaking apart on impact (Z\_12, 81 frames at $1600\times900$), all at 24 frames per second.

Each agent receives the input video with the benchmark's task description, world format, and render settings. It must deliver a build script that regenerates the 4D scene of Section~\ref{sec:4d} without reading the video. The scene contains the camera, a triangle mesh of every rendered object at every frame, persistent trajectories of the dynamic material when the solver tracks material identity, and a rendering with the frame count, resolution, and frame rate of the input video. Five harnesses converted each case once with SceneAgent skills and once without them: Gemini 4 Argon in Antigravity, GPT-6 Astra in Codex, Claude Opus 5.5 in Claude Code, Grok 4.7 in Grok CLI, and GLM-5.3 in Stirrup.

\paragraph{Metrics.}
We report the benchmark's dynamic IoU, which compares the image masks of dynamic matter in the reference, $D_t$, and in the reconstruction, $\hat D_t$, excluding a thin band $B_t$ along their boundaries,
\begin{equation}
\mathrm{DynIoU}=\frac{1}{F}\sum_{t=1}^{F}\frac{\big|(D_t\cap\hat D_t)\setminus B_t\big|}{\big|(D_t\cup\hat D_t)\setminus B_t\big|}.
\label{eq:dyniou}
\end{equation}
It ranges from 0 to 1, with higher values indicating a closer match, and we average it over the three cases.

\paragraph{Paired comparison.}
With SceneAgent, the mean dynamic IoU increases for all five harnesses (Figure~\ref{fig:4dcodebench-tokens}): from 0.86\% to 99.09\% for Gemini 4 Argon, from 11.10\% to 66.50\% for Grok 4.7, from 52.16\% to 55.95\% for Claude Opus 5.5, from 12.31\% to 39.28\% for GLM-5.3, and from 7.79\% to 36.38\% for GPT-6 Astra.

\paragraph{Comparison with published results.}
Figure~\ref{fig:4dcodebench-tokens} also shows the 18 agents in the benchmark's reference table, with token counts averaged over its 100 synthetic tasks. On these three cases, the highest mean dynamic IoU among them is 30.57\%, by Claude Opus 5.5 [High] in Claude Code. These agents were run by the benchmark authors with their own settings and harness versions.

\subsection{Qualitative Scene and Object Conversion}
\label{sec:qualitative}

\begin{figure}[t]
\centering
\includegraphics[width=\linewidth]{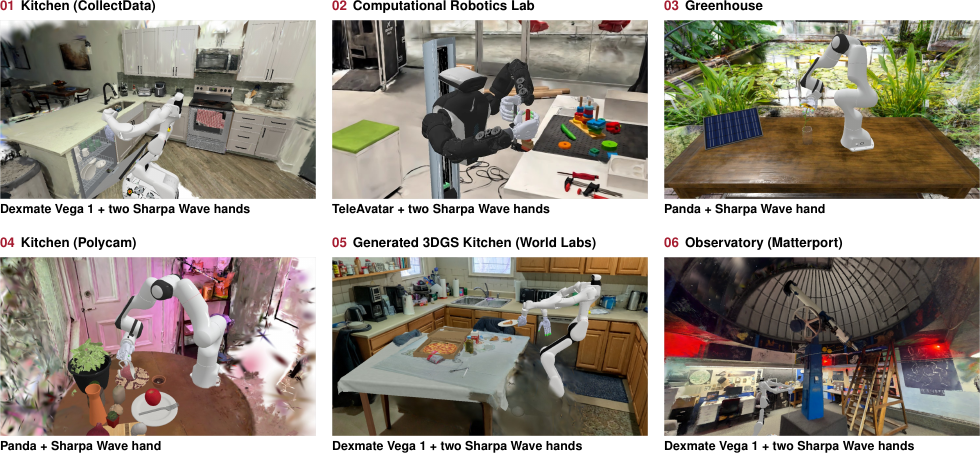}
\caption{Converted scenes shown during simulated rollouts. Captured environments and objects can be combined to construct multi-step manipulation tasks. In the Matterport observatory, a Dexmate with two Sharpa hands pulls both guide ropes to slew the telescope.}
\label{fig:scenes}
\end{figure}

\begin{figure}[t]
\centering
\includegraphics[width=\linewidth]{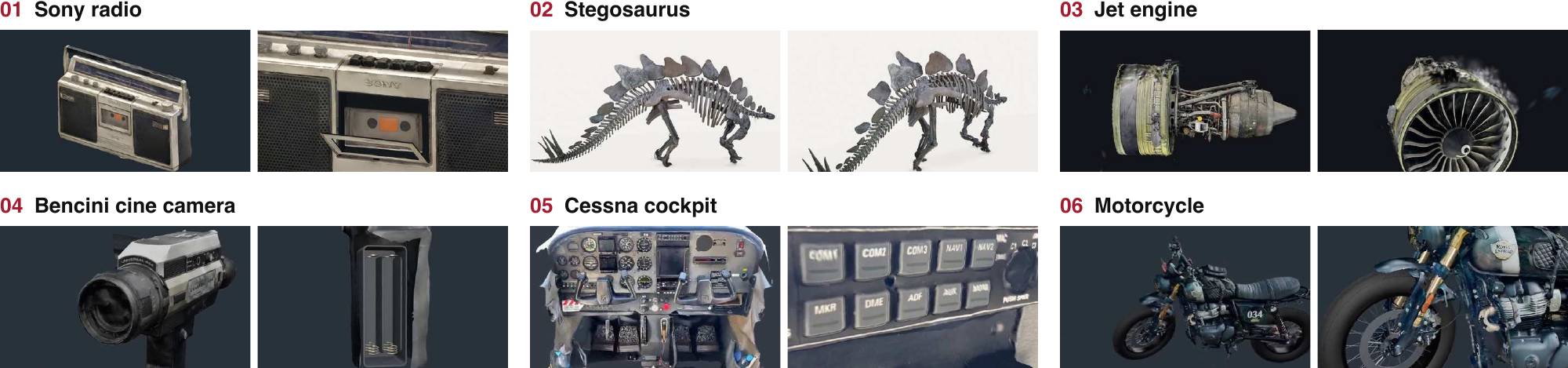}
\caption{Six objects converted from public Polycam captures, with close-up views used during agent inspection. The examples illustrate reconstruction of detailed geometry and assembly of simulation assets.}
\label{fig:objects}
\end{figure}

The converted assets illustrate scene-level reconstruction, object completion, articulation, deformation, and fracture. Figure~\ref{fig:scenes} shows six converted scenes during simulated rollouts, including a Matterport observatory capture in which a Dexmate pulls both guide ropes to slew the telescope. Figure~\ref{fig:objects} shows six converted objects with close-up inspection views. The marker, cap, and fracture examples illustrate how additional geometry and physical modeling support interactions beyond rigid pick-and-place.

We used these converted scenes to train and evaluate $\pi_{0.5}$~\cite{intelligence2025pi05} on a range of diverse tasks with a Franka Emika Panda fitted with UMI grippers. Real-world rollouts are still in progress.

\subsection{Preliminary Real-World Policy Transfer}
\label{sec:transfer}

We fine-tune pretrained policy  $\pi_{0.5}$~\cite{intelligence2025pi05} 
on demonstrations generated in reconstructed scenes of a university laboratory. The demonstration dataset used for finetuning contains different number of simulated episodes with object and scene variations. Real-world evaluation uses a Franka Emika Panda robots with UMI grippers, with a pick-and-place task. The pick-and-place task we choose to test the policies on is putting a toy eggplant into a basket. We determine a maneuver of the robot being successful as the Franka arm picks up the toy eggplant and drop it into the basket without dropping the toy eggplant in the process.

We deploy a variety of variation and randomization during demonstration generation and testing. In order for the trained model to generalize in the real-world deployment, we employ the following variations between episodes when generating successful demonstration episodes: object positions drawn randomly in a fixed bounding box with the size of $42.9 \times 9.8$ (cm), yaws of the object orientation sampled randomly in $[0, 2\pi)$, target-object size scale with $\pm 10 \%$, base position of the Franka arm offset in a 2 cm-radius disc, exterior-camera pitch varies in $\pm 2$ degrees. In the layouts of tasks during evaluation, we only employ randomization in the object positions.

All policies in the latest round are LoRA fine-tunes of the same pretrained $\pi_{0.5}$ (DROID) checkpoint, which serves as the baseline initialization. Fine-tuning applies LoRA (rank 16, $\alpha = 16$ on the Gemma-2B backbone; rank 32, $\alpha = 32$ on the 300M action expert) while the base transformer weights stay frozen; the SigLIP vision encoder and the small action/state/time projection heads train at full rank. Every policy trains for 20k steps at batch size 32 on a single H200, with each training job takes roughly 6.5 hours. And during evaluation, we run checkpoint 19999 of each policy on the tasks. We train all the policies in a simulation-only manner: all demonstrations are collected in the simulation environment and no real world demonstration is used. 

\begin{figure}[htbp]
\centering
\begin{minipage}[t]{0.49\linewidth}
\vspace{0pt}
\centering
\small
\setlength{\tabcolsep}{3pt}
\begin{tabular}{@{}rcccc@{}}
\toprule
Episodes & \multicolumn{2}{c}{Simulation} & \multicolumn{2}{c}{Real world} \\
\cmidrule(lr){2-3}\cmidrule(lr){4-5}
 & Successes & Rate & Successes & Rate \\
\midrule
10  & 9/200   & 4.5\%  & /     & / \\
20  & 13/200  & 6.5\%  & 0/50  & 0.0\% \\
40  & 57/200  & 28.5\% & 24/50 & 48.0\% \\
100 & 55/200  & 27.5\% & 20/50 & 40.0\% \\
200 & 90/200  & 45.0\% & 10/50 & 20.0\% \\
400 & 76/200  & 38.0\% & 12/50 & 24.0\% \\
1k  & 99/200  & 49.5\% & /     & / \\
2k  & 114/200 & 57.0\% & 11/50 & 22.0\% \\
4k  & 127/200 & 63.5\% & 6/50  & 12.0\% \\
10k & 102/200 & 51.0\% & /     & / \\
20k & 117/200 & 58.5\% & /     & / \\
\bottomrule
\end{tabular}
\end{minipage}\hfill
\begin{minipage}[t]{0.49\linewidth}
\vspace{0pt}
\centering
\includegraphics[width=\linewidth]{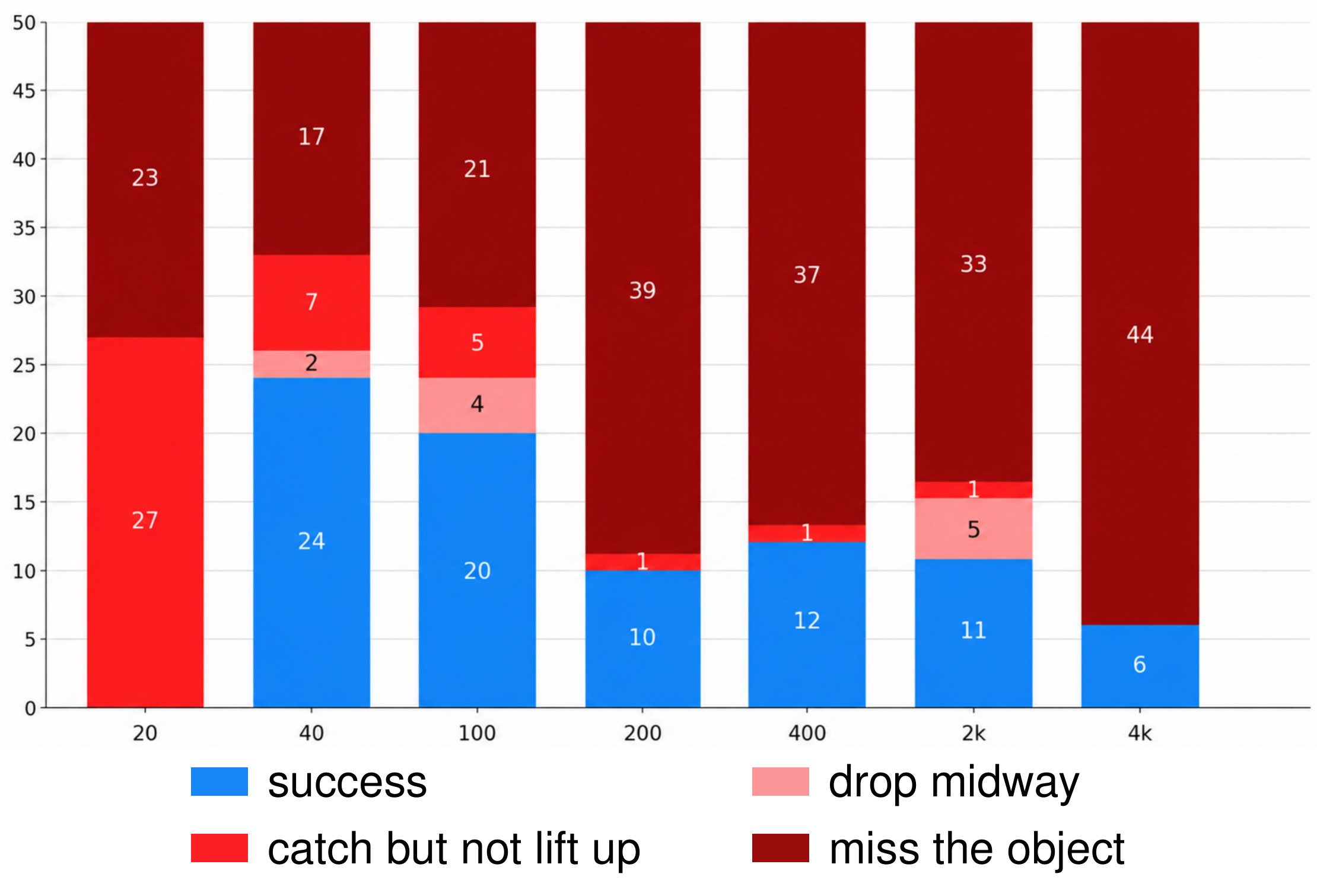}
\end{minipage}
\par\vspace{6pt}\nointerlineskip
\begin{minipage}[t]{0.49\linewidth}
\captionof{table}{Performance of policies trained on different number of demonstration episodes.}
\label{tab:scaling_comparison}
\end{minipage}\hfill
\begin{minipage}[t]{0.49\linewidth}
\captionof{figure}{Success and failure modes in real-world deployment of policies trained using different number of demonstrations.}
\label{fig:eval_failure_mode}
\end{minipage}
\end{figure}

We first present a comprehensive examination of the effect of scaling the number of simulated episodes on the real-world performance. Table ~\ref{tab:scaling_comparison} shows the success rates of different policies completing the pick-and-place task. We use 200 layouts in evaluation in simulation and use 50 layouts in real-world evaluation. Success rates in both simulation and real-world do not improve indefinitely with the scaling of demonstration dataset. The real-world success rate drops significantly after scaling past 40 demonstration episodes, indicating that excessive demonstrations alone may create severe overfitting. Figure ~\ref{fig:eval_failure_mode} further shows the different failure modes of policies in real-world deployments.

\begin{figure}[t]
\centering
\includegraphics[width=\linewidth]{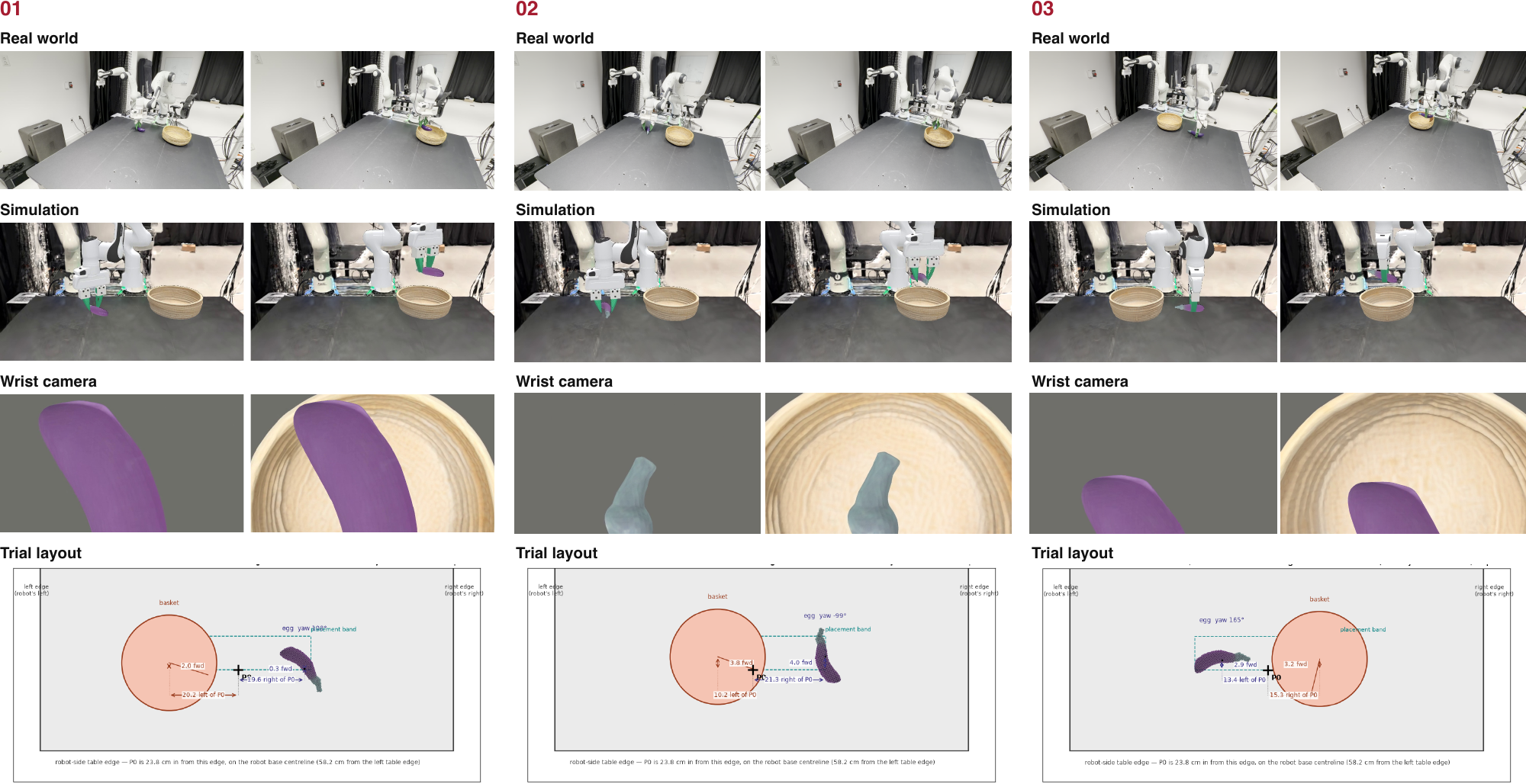}
\caption{Preliminary real-world deployment on a Franka Emika Panda after simulation-based fine-tuning. Three episodes are shown with corresponding simulated scenes and wrist-camera views, including the eggplant and basket layouts.}
\label{fig:eval}
\end{figure}

Now we present the effect of utilizing meta-learning strategies on the real-world performance. In order to achieve different meta-learning strategies, we curate three simulated dataset with different purposes. Dataset 1 consists of 2k demonstrations of the eggplant pick-and-place task with all the variations within episodes. Dataset 2 consists of 6k tasks. It is aimed at training a policy that can do three pick-and-place tasks: pick and place an eggplant, a purple cube and a cucumber. Among those 6k demonstration episodes, 2k are dedicated to each object respectively. Dataset 3 consists of 2k demonstration episodes of the eggplant pick-and-place task. But these 2k demonstration episodes do not have full variation and randomization. Among these 2k demonstrations in dataset 3, only the positions of the object are varied. With the three aforementioned curated datasets, we train three polices using dataset 1, dataset 1 $+$ dataset 2, dataset 1 $+$ dataset 2 $+$ dataset 3, respectively. Table ~\ref{tab:meta_comparison} shows that performance improves significantly when we deploy the meta learning strategy of training on dataset 1 to 3 in sequence. Figure~\ref{fig:eval} presents three real-world deployment episodes of policy 3 alongside their simulated counterparts and wrist-camera observations. 

\begin{table}[ht]
\centering
\small
\begin{tabular}{lcccccc}

\toprule
 training strategies & performance in sim & success rate in sim & real-world performance & real-world success rate \\
\midrule

 policy 1 & 114/200 & 57.0 \% &  11/50 &  22.0 \% \\
 policy 2 & 140/200 & 70.0 \% &  31/50 &  62.0 \% \\
 policy 3 & 170/200 & 85.0 \% &  44/50 &  88.0 \% \\

\bottomrule
\end{tabular}
\caption{Comparison of different training strategies. Policy 1 is trained on 2k eggplant demonstration episodes from the dataset with variations. Policy 2 is trained for 3 tasks, each task with 2k demonstration episodes with variations. Policy 3 is finetuning policy 2 on 2k eggplant demonstration episodes without variation.}
\label{tab:meta_comparison}
\end{table}

\section{Limitations and Future Work}
\label{sec:limitations}

\paragraph{Conversion cost and verification.}
Scene conversion requires repeated VLM calls and an external GPU workstation. Processing all objects in a detailed environment remains slow and error-prone. Visual inspection can identify geometric inconsistencies, but it cannot establish the accuracy of hidden geometry or physical parameters. Background completion and generated object variants also rely on predictions of content absent from the capture.

\paragraph{Physical and visual fidelity.}
Estimated mass, friction, rigidity, and articulation parameters have not yet been calibrated against measured robot interactions. The fracture examples additionally use engineering estimates. Captured Gaussian splats retain the lighting of the original scene, so relighting and compositing mesh objects can introduce inconsistent shadows and reflections. Future work should validate physical parameters against measurements and assess their effect on contact-rich manipulation, including deformation, tearing, and adhesive contact.

\paragraph{Evaluation scope.}
The reconstruction study uses one kitchen scene and one cluttered bin scene, each converted once per configuration. Broader evaluation with more scenes and repeated conversion runs is needed to characterize generalization and variability. The 4DCodeBench study covers three synthetic scenes, and real-world 4D videos, which lack 3D ground truth, remain to be evaluated. The real-world study currently focuses on pick-and-place. Object variants, placement, and lighting were randomized together, so their individual effects have not been isolated. Paired real and simulated trials are also needed to assess the predictive value of the environments for policy evaluation.

\paragraph{Further automation.}
Future directions include isolating the effects of scene and physical-parameter randomization, extending the viewer to inspect recorded trajectories, and moving capture processing and policy adaptation onto robot hardware so that data can remain on the device.

\section{Conclusion}
\label{sec:conclusion}

We presented SceneAgent, an agentic pipeline for converting 3D and 4D captures into interactive simulation environments. Its central workflow coordinates reconstruction and physical modeling tools with simulator-based inspection and repair, and for videos of dynamic scenes it fits simulated dynamics to the observed motion. Digital sisters, hybrid scenes, and an automated research loop, which turns real-robot failures into simulated demonstrations, extend the converted environments for policy training and evaluation.

On a cluttered bin scene from GraspClutter6D, SceneAgent raises F1 at 1~cm for every harness with scorable paired outputs, and its best F1 score is 60.75\%, compared with 6.9\% for SimFoundry. On three dynamic scenes from 4DCodeBench, SceneAgent raises the mean dynamic IoU of Gemini 4 Argon from 0.86\% to 99.09\%. Qualitative examples demonstrate a range of simulated interactions and preliminary real-world deployment after policy fine-tuning. Quantitative policy evaluation and validation of estimated physical parameters remain ongoing. The code, processed scenes, and example converted objects will be released with the paper.

\nocite{shen20264dcodebench}
\bibliographystyle{plainnat}
\bibliography{references}

\section*{Featured Art and 3D Captures}

{\small\renewcommand{\arraystretch}{1.15}
\begin{tabularx}{\linewidth}{@{}>{\raggedright\arraybackslash}Xl@{}}
\toprule
Asset & Creator \\
\midrule
\multicolumn{2}{@{}l}{\textbf{Sketchfab}} \\
\href{https://sketchfab.com/3d-models/rudbeckia-flower-eee1a1ebdb614f38a722b52e26ebe038}{Rudbeckia flower} & 3dhdscan \\
\href{https://sketchfab.com/3d-models/chinese-statue-e0ccdb444522470f9e6567870e63c40c}{Chinese statue} & netgis \\
\href{https://sketchfab.com/3d-models/asus-z170-p-motherboard-b998596cfc4945a0bc7b016005c39321}{Asus Z170-P motherboard} & Lassi Kaukonen \\
\href{https://sketchfab.com/3d-models/rx-480-gpu-61e69f50bcd44c7284bef31a8f21c6a7}{RX 480 GPU} & Igor.Jop \\
\addlinespace[4pt]
\multicolumn{2}{@{}l}{\textbf{Polycam}} \\
\href{https://poly.cam/capture/0072f705-655d-4860-8468-2f12968e503c}{Vintage Sony radio CFS-43} & JordanFry \\
\href{https://poly.cam/capture/cae4a43a-87aa-4ed8-9e21-cff222662d0b}{Totoro} & TechnologicalFrostfang2470 \\
\href{https://poly.cam/capture/BB04FF30-6D58-4BE8-8706-3227D1FD8EBF}{Stegosaurus composite skeleton, American Museum of Natural History} & MTP \\
\href{https://poly.cam/capture/C80F9BB5-E4DE-4466-9018-25C1F271C250}{Bencini Comet Universal 444 camera (1977)} & redline3678 \\
\href{https://poly.cam/capture/5DF483F7-5ECB-48EA-BDFC-EA4FD93A5A2C}{Cessna C172R cockpit} & wlanfr3ak \\
\href{https://poly.cam/capture/7B46071B-A067-4ACF-A4EE-DC568F65217F}{Motorcycle} & ZanyFalcon4404 \\
\href{https://poly.cam/capture/BB7E52C7-BE42-4F57-9839-0969215A6C50}{Towel rack} & Pandora \\
\href{https://poly.cam/capture/0E267738-A32F-4160-8AE6-A55A804B87CC}{Red towel} & AutomatedStarfish942 \\
\href{https://poly.cam/capture/973DB233-4029-4CB4-8A15-22BF483C42CB}{Dept Artist\'e anorak} & Artist \\
\href{https://poly.cam/capture/158AE9E1-ED96-4661-B02D-8308B7F576FA}{Firefighter hose} & drewjosh \\
\href{https://poly.cam/capture/0EB1FB71-49AE-4FC3-B2A5-785AF793EFE0}{Electric extension lead} & sisamm \\
\href{https://poly.cam/capture/28520EB0-CAFB-400C-B93C-6BB9D3795262}{Liquid Death hat} & stevieburd \\
\href{https://poly.cam/capture/2C13178D-68EE-4343-A0C6-02CB1E238492}{Black-figure Dinos and stand with the wedding of Peleus and Thetis, British Museum} &  \\
\href{https://poly.cam/capture/1ABBF807-C43B-4531-83E5-AA74D22B8E61}{Small Earthenware Pot: EOA:2022.111.18} & PortlandMuseum \\
\bottomrule
\end{tabularx}}

\end{document}